\documentclass[letterpaper]{article} 
\usepackage{aaai24}  
\usepackage{times}  
\usepackage{helvet}  
\usepackage{courier}  
\usepackage[hyphens]{url}  
\usepackage{graphicx} 
\usepackage{natbib}  
\usepackage{caption} 
\usepackage{algorithm}
\usepackage{algorithmic}

\usepackage{newfloat}
\usepackage{listings}

\usepackage{amsfonts}       
\usepackage{amsmath,amssymb}
\usepackage{nicefrac}       
\usepackage{microtype}      
\usepackage{xcolor}         
\usepackage{subfigure}
\usepackage{tabularx}
\usepackage{multirow}
\usepackage{rotating}
\usepackage{makecell}
\usepackage{booktabs}
\usepackage[shortlabels]{enumitem}
\usepackage{amsthm}   
\usepackage{algorithm,algorithmic}
\usepackage{mathtools}
\usepackage{diagbox}
\usepackage{makecell}
\usepackage[shortlabels]{enumitem}
\newtheorem{theorem}{Theorem}
\newtheorem{assumption}{Assumption}

\newtheorem{challenge}{Challenge}
\newtheorem{principle}{Principle}

\newcommand{\bo}[1]{\textbf{#1}}
\newcommand{\app}{Snowball }
\newcommand{\appnoblank}{Snowball}
\DeclareCaptionStyle{ruled}{labelfont=normalfont,labelsep=colon,strut=off} 
\floatstyle{ruled}
\newfloat{listing}{tb}{lst}{}
\floatname{listing}{Listing}
\title{Resisting Backdoor Attacks in Federated Learning via Bidirectional Elections \\ and Individual Perspective}
\author {
    Zhen Qin\textsuperscript{\rm 1},
    Feiyi Chen\textsuperscript{\rm 1},
    Chen Zhi\textsuperscript{\rm 2},
    Xueqiang Yan\textsuperscript{\rm 3},
    Shuiguang Deng\textsuperscript{\rm 1}\thanks{Shuiguang Deng is the corresponding author.}
}
\affiliations {
    \textsuperscript{\rm 1}College of Computer Science and Technology, Zhejiang University, Hangzhou, China\\
    \textsuperscript{\rm 2}School of Software Technology, Zhejiang University, Ningbo, China\\
    \textsuperscript{\rm 3}Huawei Technologies Co. Ltd., Shanghai, China\\
    \{zhenqin, chenfeiyi, zjuzhichen\}@zju.edu.cn, yanxueqiang1@huawei.com, dengsg@zju.edu.cn
}

\begin{document}

\maketitle

\begin{abstract}
Existing approaches defend against backdoor attacks in federated learning (FL) mainly through a) mitigating the impact of infected models, or b) excluding infected models.
The former negatively impacts model accuracy, while the latter usually relies on globally clear boundaries between benign and infected model updates.
However, in reality, model updates can easily become mixed and scattered throughout due to the diverse distributions of local data.
This work focuses on excluding infected models in FL.
Unlike previous perspectives from a global view, we propose \appnoblank, a novel anti-backdoor FL framework through bidirectional elections from an individual perspective inspired by one principle deduced by us and two principles in FL and deep learning.
It is characterized by a) bottom-up election, where each candidate model update votes to several peer ones such that a few model updates are elected as selectees for aggregation;
and b) top-down election, where selectees progressively enlarge themselves through picking up from the candidates.
We compare \app with state-of-the-art defenses to backdoor attacks in FL on five real-world datasets, demonstrating its superior resistance to backdoor attacks and slight impact on the accuracy of the global model.
\end{abstract}

\section{Introduction}
\label{sec-intro}
Federated Learning (FL) \cite{mcmahan2017fl} enables multiple devices to jointly train machine learning models without sharing their raw data. 
Due to the unreachability to distributed data, it is vulnerable to attacks from malicious clients \cite{wang2020attack-tails}, especially \emph{backdoor attacks} that neither significantly alter the statistical characteristics of models as Gaussian-noise attacks \cite{blanchard2017machine-Krum} nor cause a distinct modification to the training data as label-flipping attacks \cite{liu2021privacy}, and thus, are more covert against many existing defenses \cite{zeng2022never-too-late}.

\textbf{Existing defenses} to backdoor attacks in FL are mainly based on a) mitigating the impact of infected models
\cite{bagdasaryan2020backdoor-howto,sun2019can,xie2021crfl,nguyen2021-flame,zhang2023flip} or b) excluding infected models based on their deviations \cite{blanchard2017machine-Krum,ozdayi2021defending-RLR,fung2018mitigating,Rieger2022deepsight,li2020learning,shejwalkar2021manipulating,zhang2022fldetector,shi2022-FAA-DL,nguyen2021-flame}. 
The former may negatively impact global model accuracy \cite{DBLP:conf/iclr/YuZ0L21}.
The latter assumes globally clear boundaries between benign and infected model updates \cite{zeng2022never-too-late}.
However, backdoor attacks typically manipulate a limited subset of parameters, resulting in the similarity between benign and infected model updates.
Besides, the nature of Non-Independent and Identically Distributed (non-IID) data in FL increases diversity among model updates.

\begin{figure}[t]
  \centering
  \includegraphics[width=0.50\linewidth]{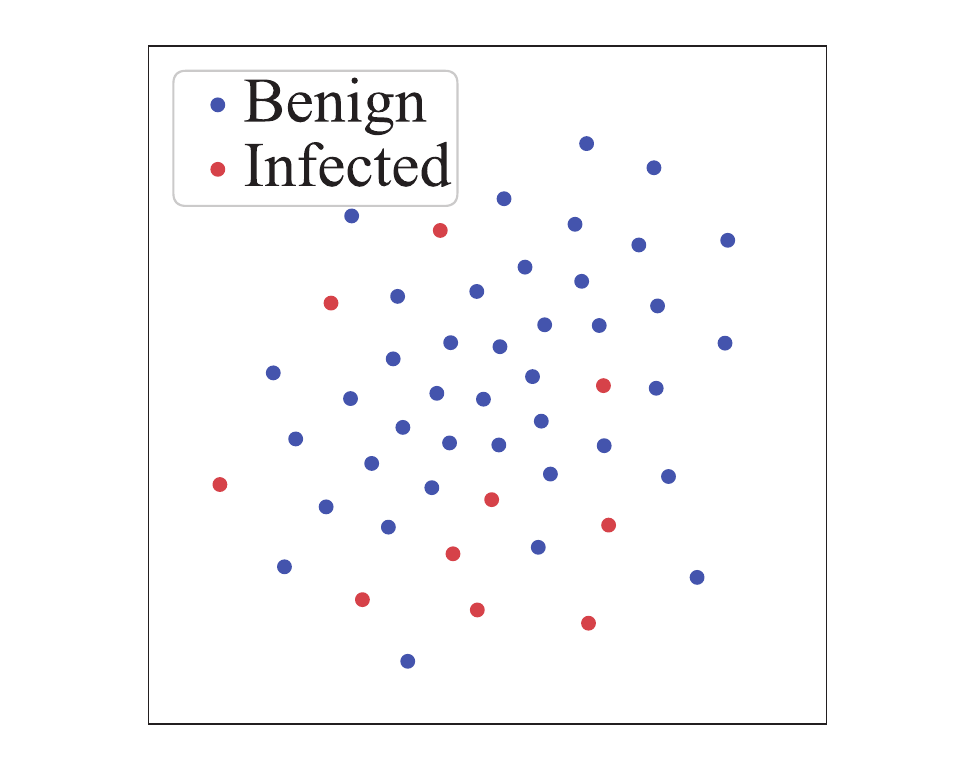}
  \caption{2D-visualized 50 model updates in one round of FL (practical non-IID MNIST with $\alpha$=0.5, PDR=0.3).}
  \label{pic-intuition-mnist-scatter}
\end{figure}
Actually, benign and infected model updates are easy to be mixed with complicatedly non-IID data 
(practical non-IID \cite{hsu2019measuring,huang2021-FedAMP} and feature distribution skew \cite{tan2022-PFL-survey}),
or with not very high poison data ratio (PDR). 
We experimentally demonstrate it in Figure \ref{pic-intuition-mnist-scatter}, where benign and infected updates are mixedly scattered. 
In such cases, anomaly detections based on linear similarity may not perform satisfactorily.
Besides, when facing a relatively high malicious client ratio (MCR), infected model updates are easier to be mistreated as benign ones, however, many existing defenses are only evaluated with MCR $\leq$ 10\% \cite{xie2021crfl,ozdayi2021defending-RLR,zeng2022never-too-late,lu2022defense}.
Although model deviations may be better captured by nonlinear neural networks, \emph{the patterns of benign models in FL are usually hard to acquire} due to unpredictable distributions and trajectory shifts of model updates. \citet{li2020learning} use the test data to generate model weights for training the detection model, but the test data with a similar distribution to all clients may be usually unavailable. 
Besides, model weights usually follow extremely complex distributions, making them hard to learn.

To better leverage powerful neural networks to detect malicious models, we propose \appnoblank, an anti-backdoor FL framework taking advantage of linear and non-linear approaches, i.e., without the need for pre-defined benign patterns and the powerful capability to capture model deviations, respectively. 
It treats each model update as an agent electing model updates for aggregation with an individual perspective, where the motivation comes from that: \textbf{defenses of the models, by the models, for the models.} From a global perspective as existing studies \cite{nguyen2021-flame,ozdayi2021defending-RLR,blanchard2017machine-Krum,li2020learning}, benign and infected model updates may appear mixed.
If we examine model updates from the perspective of individual model updates, the nearest ones may have the same purpose since both benign and infected updates wish to exclude each other from aggregation. Thus, if we make each model vote for the closest model updates, the benign model updates may get more votes when the benign clients account for the majority. 

The elections in \app are bidirectional and conducted sequentially, i.e., 1) bottom-up election where candidate model updates nominate a small group of peers as selectees to be aggregated; and 2) top-down election that regards the selectees as benign patterns and progressively enlarges the number of selectees from the rest candidates through a variational auto-encoder (VAE), which focuses on the model-wise differences instead of benign patterns themselves.
We know it may be difficult to have a one-size-fits-all approach, fortunately, \app can be easily integrated into existing FL systems in a non-invasive manner, since it only filters out several model updates for aggregation.
For attacks that have not been mentioned in this work, aggregation can be conducted on the intersection between the updates selected by existing approaches and that of \appnoblank.

The main contributions of this work lie in:
\begin{enumerate}[itemindent=0em, listparindent=0em, leftmargin=1em]
  \item Proposing a novel anti-backdoor FL framework named \appnoblank. It selects model updates with bidirectional elections from an individual perspective, contributing to the leverage of neural networks for infected model detection. 
  \item Proposing a new paradigm for utilizing VAE to detect infected models, i.e., progressively enlarges the selectees with focusing on the model-wise differences instead of benign patterns themselves, to better distinguish infected model updates from benign ones.
  \item Conducting extensive experiments on 5 real-world datasets to demonstrate the superior attack-resistance of \app over state-of-the-art (SOTA) defenses when the data are complicatedly non-IID, PDR is not very high and the ratio of attackers to all clients is relatively high.
  Also, \app brings a slight impact on the global model accuracy. Codes are available at https://github.com/zhenqincn/Snowball.
\end{enumerate}

\section{Related Work}
\label{sec-related-work}
Existing work defends targeted attacks in FL by 
a) mitigating the impact of infected models, including
a1) robust learning rate \cite{ozdayi2021defending-RLR,fung2018mitigating}, 
a2) provably secure FL by model ensemble \cite{xie2021crfl,cao2021provably},
a3) adversarial learning \cite{zhang2023flip},
or 
b) filtering out infected models or parameters, including:
b1) Byzantine-robust aggregation \cite{blanchard2017machine-Krum,yin2018byzantine-trimmed-mean}, and
b2) anomaly detection \cite{li2020learning,zhang2022fldetector,shi2022-FAA-DL,shejwalkar2021manipulating,zhang2022fldetector}.  
Besides, there are also approaches that combine weight-clipping, noise-addition and clustering \cite{bagdasaryan2020backdoor-howto,sun2019can,nguyen2021-flame,Rieger2022deepsight}, which belong to both of the two main categories.

These approaches are validated to be effective in different scenarios. 
However, approaches mitigating the impact of infected models usually lower the global model accuracy (a1, a2) or rely on certain assumptions which may not be always satisfied and cause inference latency and memory consumption (a3) \cite{li2022backdoor-survey}. 
Approaches filtering out infected models usually require globally clear boundaries between benign and infected model updates \cite{zeng2022never-too-late}, which usually only occur when 1) the non-IIDness of data is not complex (IID or pathological non-IID) where model updates are easy to form distinct clusters \cite{nguyen2021-flame,ozdayi2021defending-RLR,Rieger2022deepsight} or 2) the PDR is high ($\geq$ 50\%) such that infected model updates deviate significantly from benign ones \cite{Rieger2022deepsight,ozdayi2021defending-RLR}.
Besides, many defenses are only evaluated with MCR $\leq$ 10\% \cite{xie2021crfl,ozdayi2021defending-RLR,zeng2022never-too-late,lu2022defense}.

Thus, there is a strong demand for an approach that can effectively defend against backdoor attacks when benign and infected models are scattered without clear boundaries.

\section{Background}
This work focuses on the classical FL \cite{mcmahan2017fl}. 
Let $\mathbb{D} = \left\{\mathcal{D}_1, \mathcal{D}_2, \ldots, \mathcal{D}_N \right\}$ denote the datasets held by the $N$ clients respectively. The goal of FL is formulated as:
\begin{equation}
  \min_{\mathbf{w}} f(\mathbf{w}) \coloneqq \sum_{i=1}^N \lambda_i f(\mathbf{w}, \mathcal{D}_i)
  \label{eq-fl-objective}
\end{equation}
where $f(\mathbf{w}, \mathcal{D}_i) \!\!\coloneqq\!\! \frac{1}{|\mathcal{D}_i|} \sum_{\xi \in \mathcal{D}_i, \xi \sim \mathcal{Z}_i} \ell (\mathbf{w}, \xi)$ is the average loss $\ell$ on data sample $\xi$ of client $i$, where $\xi$ follows distribution $\mathcal{Z}_i$, and $\lambda_i$ is the weight of client $i$. 
In each round $t$ of the total $T$ rounds, $K$ ($K \!\leq\! N$) clients are randomly selected as participants.
Participant $i$ trains $\mathbf{w}$ to minimize $f$ for $E$ epochs and submit its model update $\Delta \mathbf{w}_{i,t}$ to the server for aggregation.
A certain proportion of the participants in each round conduct backdoor attacks, referred to as \emph{attackers}. 

\section{Methodology}
\subsection{Overview}
\label{subsec-approach-overview}
\begin{figure*}[t]
  \centering
  \includegraphics[width=0.78\linewidth]{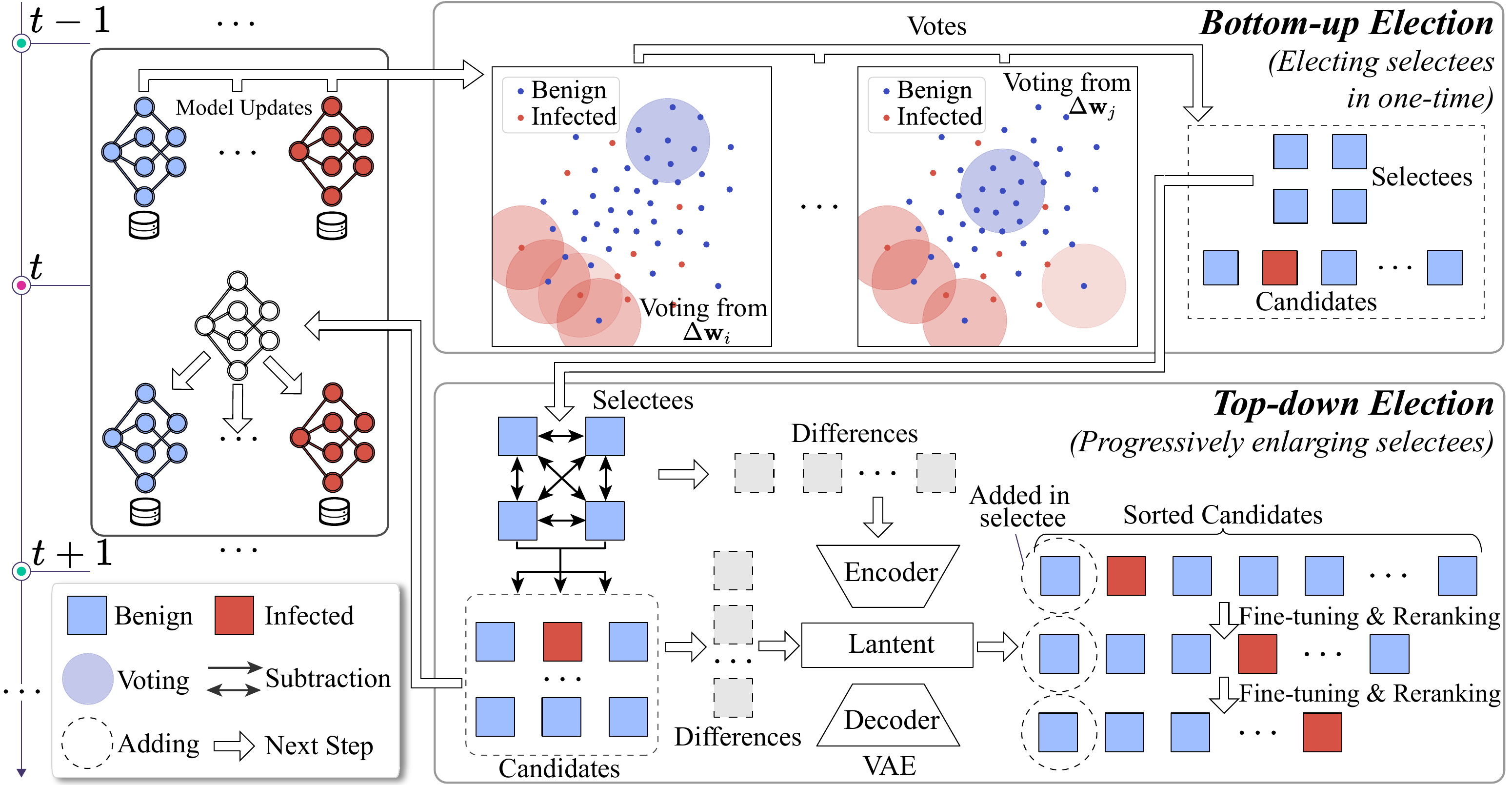}
  \caption{Overview of \appnoblank, which improves the aggregation procedure in FL on the server.}
  \label{pic-overview}
\end{figure*}
Designing an anti-backdoor approach based on anomaly detection may better preserve the accuracy of the global model since no noise is introduced.
However, there are two main challenges in adopting anomaly detection techniques:
\begin{challenge}[Insufficient Benign Pattern]
  Due to unpredictable distributions and trajectory shifts of model updates, there lacks patterns for benign model updates in each round.
  \label{challenge-no-normal-pattern}
\end{challenge}
\begin{challenge}[Ambiguous Boundary]
  The boundary between benign and infected model updates is usually unclear due to the mild impact of backdoor attacks on model parameters and the non-IIDness of FL.
  \label{challenge-model-mixed}
\end{challenge}

To address these challenges, \app goes through two election procedures sequentially before aggregation in each round, i.e., \emph{bottom-up election} and \emph{top-down election}, as shown in Figure \ref{pic-overview}.
\emph{Bottom-up election} is designed with the inspiration of \cite{shayan2021biscotti,qin2023blockdfl} which shifts the view from a global perspective to an individual model perspective.
It takes the $K$ collected model updates $\mathcal{W}_t = \left\{\Delta \mathbf{w}_{i,t}\right\}^{i \in \mathcal{C}_t}$ from clients $\mathcal{C}_t$ participating round $t$ as the input, and locates a few model updates the least likely to be infected (Challenge \ref{challenge-no-normal-pattern}). 
In it, each model update votes for several ones closest to it, and a few model updates with the most votes are designated as \emph{Selectees}, denoted by $\widetilde{\mathcal{W}}_t  \subset \mathcal{W}_t$. 
Such an individual perspective helps to separate benign and infected model updates at a finer granularity (Challenge \ref{challenge-model-mixed}). 

Then, \emph{top-down election} enlarges selectees to aggregate more model updates with those in $\widetilde{\mathcal{W}}_t $ as benign patterns. 
A variational auto-encoder (VAE) \cite{kingma2013auto-VAE} is adopted to mine benign ones from $\mathcal{W}_t - \widetilde{\mathcal{W}}_t $ focusing on the differences of model updates. 
On one hand, learning the differences quadratically augments the benign patterns (Challenge \ref{challenge-no-normal-pattern}). 
On the other hand, compared with model updates, the differences among them are easier to be distinguished and learned (Challenge \ref{challenge-model-mixed}).
This process progressively enlarges selectees to continually enlarge the benign patterns.
The process of \app is described in Algorithm \ref{algo:main_algo}.

\subsection{Bottom-up Election}
\label{subsec-approach-clustering}
We will first introduce the principle behind this procedure.
\begin{principle}
  The difference between two model updates is expected to be positively correlated with the difference between their corresponding data distributions.
  \label{principle-difference-data-model}
\end{principle}
Principle \ref{principle-difference-data-model} is mentioned in many studies on the non-IIDness of FL \cite{zhao2018federated-non-IID,fallah2020personalized} and widely used by clustering-based FL \cite{ghosh2020efficient,sattler2020clustered}.
\begin{assumption}
  There is a standard distribution $\mathcal{Z}$ such that $ \forall \mathcal{Z}_i$ can be modeled as $\mathcal{Z}_i = \mathcal{Z} + \epsilon_i$, where $\epsilon_i$ is an offset of $\mathcal{Z}_i$ relative to the standard distribution.
  \label{assumption-standard-benign}
\end{assumption}
\begin{assumption}
  Injected data on client $i$ can be generated by sampling from distribution $\mathcal{Z}_i + \delta_i$, where $\delta_i$ is an offset that shifts $\mathcal{Z}_i$ to a backdoored data distribution.
  \label{assumption-standard-injected}
\end{assumption}
Assumption \ref{assumption-standard-benign} indicates that the difference between data distributions of benign clients $i$ and $j$ depends on $\epsilon_i$ and $\epsilon_j$. 
When data among clients are IID, $\forall \epsilon$ is a zero distribution.
Taken as a whole, benign and infected model updates may not be clearly distinguishable due to the diversity of $\epsilon$.
But if Assumptions \ref{assumption-standard-benign} and \ref{assumption-standard-injected} hold, those model updates closer to a benign one are more likely to be benign, as illustrated in Figure \ref{pic-intuition-mnist-scatter}.
Thus, each model update votes for those closest to them. 
More supports of Principle \ref{principle-difference-data-model} are in Appendix \ref{subsec-support-principle-difference-data-model}.

\begin{algorithm}[t]
  \caption{\textbf{Main Process of \appnoblank.}}
  \begin{algorithmic}[1]
  \STATE \textbf{Input:} Updates $\mathcal{W}$, target \# of updates in the two procedures $\check{M}$ and $M$, \# of clusters $\check{K}$ for voting, \# of epochs for training and tuning VAE $E^{VI}$ and $E^{VT}$, \# of updates added in one step of top-down election $M^E$, current round $t$, and the round to start top-down election $T^V$.
  \STATE $\widetilde{\mathcal{W}}$ = \textbf{BottomUpElection}($\mathcal{W}$, $\check{M}$, $\check{K}$)
  \STATE \textbf{if}$(t > T^V)$, $\widetilde{\mathcal{W}}$ = \textbf{TopDownElection}($\widetilde{\mathcal{W}}$, $\mathcal{W}$, $E^{VI}$, $E^{VT}$, $M$) \textbf{end if}
  \label{algo-main-line-progressive-start}
  \STATE \textbf{return} $\widetilde{\mathcal{W}}_t $
  
  \hspace{-2em}
  \underline{\textbf{BottomUpElection}($\mathcal{W}$, $\check{M}$, $\check{K}$):}
  \STATE Initial counter $c$ with zeros, where $c_i$ is for $\Delta \mathbf{w}_i$
  \FOR{layer $m = 0, 1, \ldots, L$}
    \FOR{$\mathbf{w}_i \in \mathcal{W}$}
      \STATE Select $\Delta \mathbf{w}_{j,m}$ with larger $\|\Delta \mathbf{w}_{i,m} - \Delta \mathbf{w}_{j,m}\|$ to constitute $\mathcal{W}^C_m$, where $|\mathcal{W}^C_m| = \check{K} - 1$
      \STATE $\mathbf{r}_i = $ \texttt{K-means}($\mathcal{W}$, $\mathcal{W}^C_m \cup \left\{\mathbf{w}_i - \mathbf{w}_i\right\}$, $\check{K}$), $s_i$ = \texttt{CH\_Score}($\mathbf{r}_i$) $\backslash \backslash$ clustering result and score
    \ENDFOR
    \STATE $s = $ \texttt{Min-MaxNormalization}($s$)
    \label{algo-main-line-norm}
    \STATE \textbf{for} $\Delta \mathbf{w}_i \in \mathcal{W}$ \textbf{do} \ if $r_{i,i} = r_{i,j}$ then $c_j = c_j + s_i$, $\forall \Delta \mathbf{w}_j \in \mathcal{W}$ \ \textbf{end for} 
  \ENDFOR
  \STATE \textbf{return} $\widetilde{\mathcal{W}}$ containing $\check{M}$ model updates with larger $c_i$ 

  \hspace{-2em}
  \underline{\textbf{TopDownElection}($\widetilde{\mathcal{W}}$, $\mathcal{W}$, $E^{VI}$, $E^{VT}$, $M^E$, $M$):}
  \STATE Build $\mathcal{U} \!=\! \left\{\mathbf{u}_{i,j}, \ldots \right\}$, $\mathbf{u}_{i,j} = \Delta \mathbf{w}_i - \Delta \mathbf{w}_j$, $\forall \Delta \mathbf{w}_i$, $\Delta \mathbf{w}_j \in \widetilde{\mathcal{W}} $, $i \neq j$, then train $\mathbf{v}$ for $E^{VI}$ epochs
  \label{algo-main-line-build-U}
  \WHILE {$|\widetilde{\mathcal{W}}|$ < $M$}
  \STATE Rebuild $\mathcal{U}$ as Line \ref{algo-main-line-build-U}, tune $\mathbf{v}$ on $\mathcal{U}$ for $E^{VT}$ epochs
  \STATE Select $\Delta \mathbf{w}_{j}$ with larger $\sum_{\Delta \mathbf{w}_{i} \in \widetilde{\mathcal{W}}} \operatorname{recon}(\Delta \mathbf{w}_i - \Delta \mathbf{w}_j,\mathbf{v}(\Delta \mathbf{w}_i - \Delta \mathbf{w}_j))$  from $\mathcal{W} - \widetilde{\mathcal{W}}$, denoted by $\mathcal{W}^A$ ($|\mathcal{W}^A| = M^E$), then $\widetilde{\mathcal{W}} = \widetilde{\mathcal{W}} \cup \mathcal{W}^A$  \ $\backslash \backslash$ enlarging $\widetilde{\mathcal{W}}$
  \ENDWHILE
  \STATE \textbf{return} $\widetilde{\mathcal{W}}$
  \end{algorithmic}
  \label{algo:main_algo}
\end{algorithm}

It is hard to clearly define ``closeness'', so each model update runs K-means independently to guide its voting.
For $\Delta \mathbf{w}_{i, t}$, we select $\check{K} - 1$ model updates from $\mathcal{W}_t - \left\{\mathbf{w}_{i, t}\right\}$ with the largest $\|\Delta \mathbf{w}_{i, t} - \Delta \mathbf{w}_{j, t}\|^2$ as the initial centroids of K-means together with a zero vector with the same shape as $\Delta \mathbf{w}_{i, t}$.
During implementation, $\check{K}$ is predetermined through Gap statistic \cite{tibshirani2001estimating-gap} on model updates collected in the first round, where the details can refer to in Appendix \ref{subsec-appendix-hyper-app}.
After clustering, $\check{K}$ clusters are obtained, and $\Delta \mathbf{w}_{i, t}$, as well as the model updates that belong to the same cluster as its, are voted, as shown in the upper right part of Figure \ref{pic-overview}.

We weight the clustering result from each update by Calinski and Harabasz score \cite{calinski1974dendrite} (the higher, the better) due to the sensitivity of K-means to initial centroids.
Since different layers have different parameter counts, the voting is layer-wisely conducted for $L$ times with an $L$-layer network.
The voting weights in each layer are scaled in [0, 1] by min-max normalization and then accumulated. 
Finally, $\check{M}$ updates with the highest votes form $\widetilde{\mathcal{W}}_t $.

\subsection{Top-down Election}
\label{subsec-approach-vae}
Bottom-up election provides several trusted model updates. 
However, since benign and infected updates share certain similarities, K-means, an approach relying on linear distance, cannot deeply mine their differences. 
To ensure infected updates are excluded, $\check{M}$ has to be small.
To avoid too few model updates included in aggregation such that the convergence of FL is negatively impacted, a VAE \cite{an2015variational} is introduced to learn the patterns of benign model updates by utilizing its nonlinear latent feature representation.
Although $\widetilde{\mathcal{W}}_t $ provides a few benign patterns, it is still hard to train a VAE since 1) $|\widetilde{\mathcal{W}}_t |$ is too small, and 2) samples in $\widetilde{\mathcal{W}}_t $ follow different distributions, causing large reconstruction error.
Thus, we focus on the differences between model updates rather than model updates themselves. 

\begin{principle}
  It is easier to push a stack of nonlinear layers towards zero than towards identity mapping \cite{he2016deep-ResNet}.
  \label{principle-resnet}
\end{principle}
Principle \ref{principle-resnet} is a key basis of Deep Residual Networks (ResNet) \cite{he2016deep-ResNet}. 
Let $\Delta \mathbf{w}_i^B$ and $\Delta \mathbf{w}_i^*$ denote arbitrary benign and infected model update, respectively:
\begin{principle}
  If $\Delta \mathbf{w}_i^*$ is always filtered out in each round, the difference between $\Delta \mathbf{w}_i^B$ and $\Delta \mathbf{w}_j$ is expected to have a smaller $L_2$ norm than that between $\Delta \mathbf{w}_i^*$ and $\Delta \mathbf{w}_j$ as the global model converges.
  \label{principle-ddif-trend}
\end{principle}
\noindent Principle \ref{principle-ddif-trend} is supported based on the following assumptions.
\begin{assumption}
  $\exists t^{B} \!< \!T$ such that after round $t^{B}$, $\mathbb{E}(\|\Delta \mathbf{w}_i^B \!-\! \Delta \mathbf{w}_j^B\|^2) \!-\! \mathbb{E}(\|\Delta \mathbf{w}_i^B \!-\! \Delta \mathbf{w}_j^*\|^2) \!<\! 0$.
  \label{assumption-a-b}
\end{assumption}
\begin{assumption}
  If infected updates are continually filtered out, $\exists t^C < T$ such that after round $t^C$, we have
  \begin{equation}
    \mathbb{E}(\|\Delta \mathbf{w}_i^B - \Delta \mathbf{w}_j^B\|^2) - \mathbb{E}(\|\Delta \mathbf{w}_i^* - \Delta \mathbf{w}_j^*\|^2) < 0.
  \end{equation}
  \label{assumption-a-c}
\end{assumption}

\begin{figure}[t]
  \centering
  \subfigure[MNIST (Label Skew)]{
    \includegraphics[width=0.4743\linewidth]{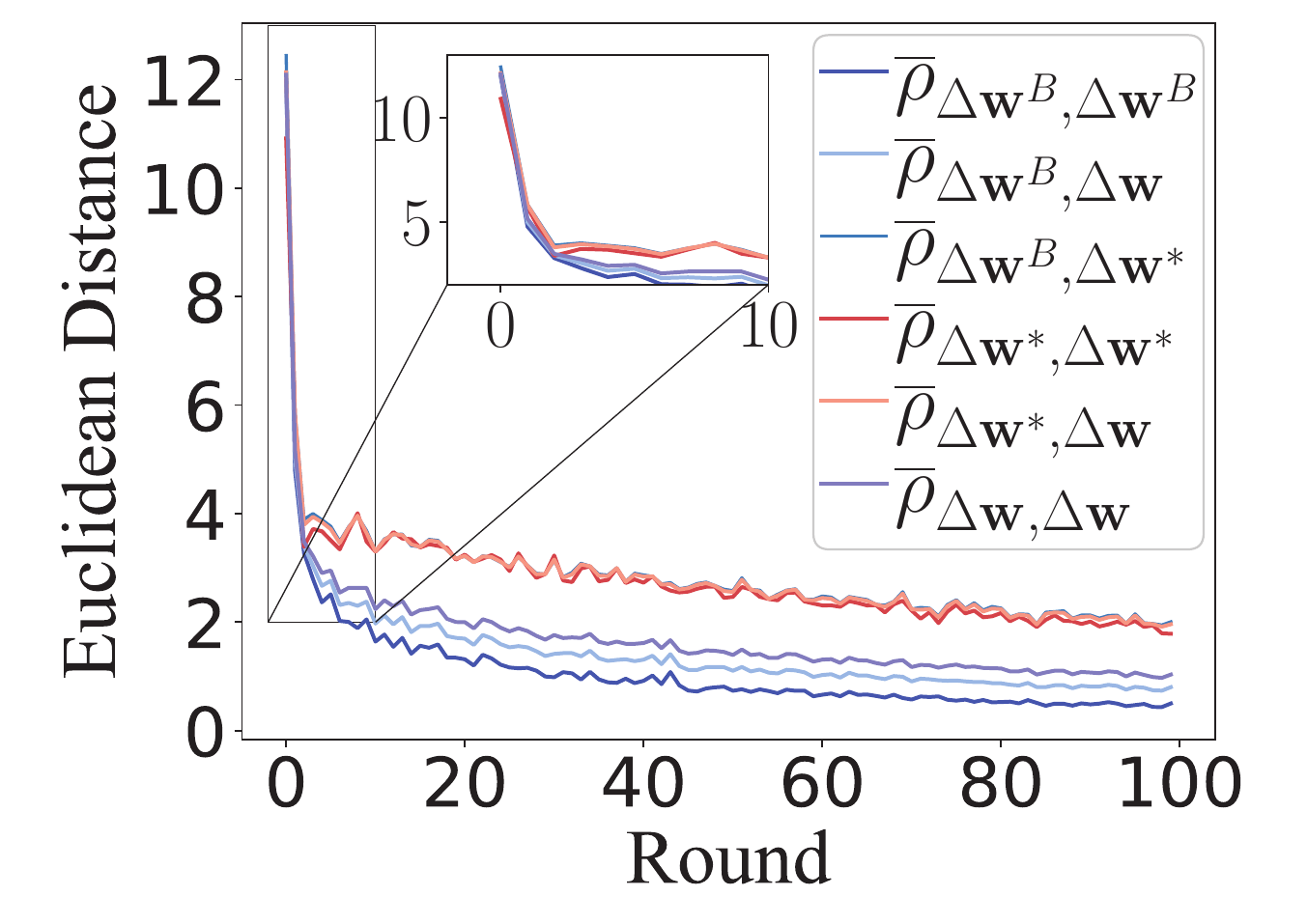}
  }
  \subfigure[FEMNIST (Feature Skew)]{
    \includegraphics[width=0.4743\linewidth]{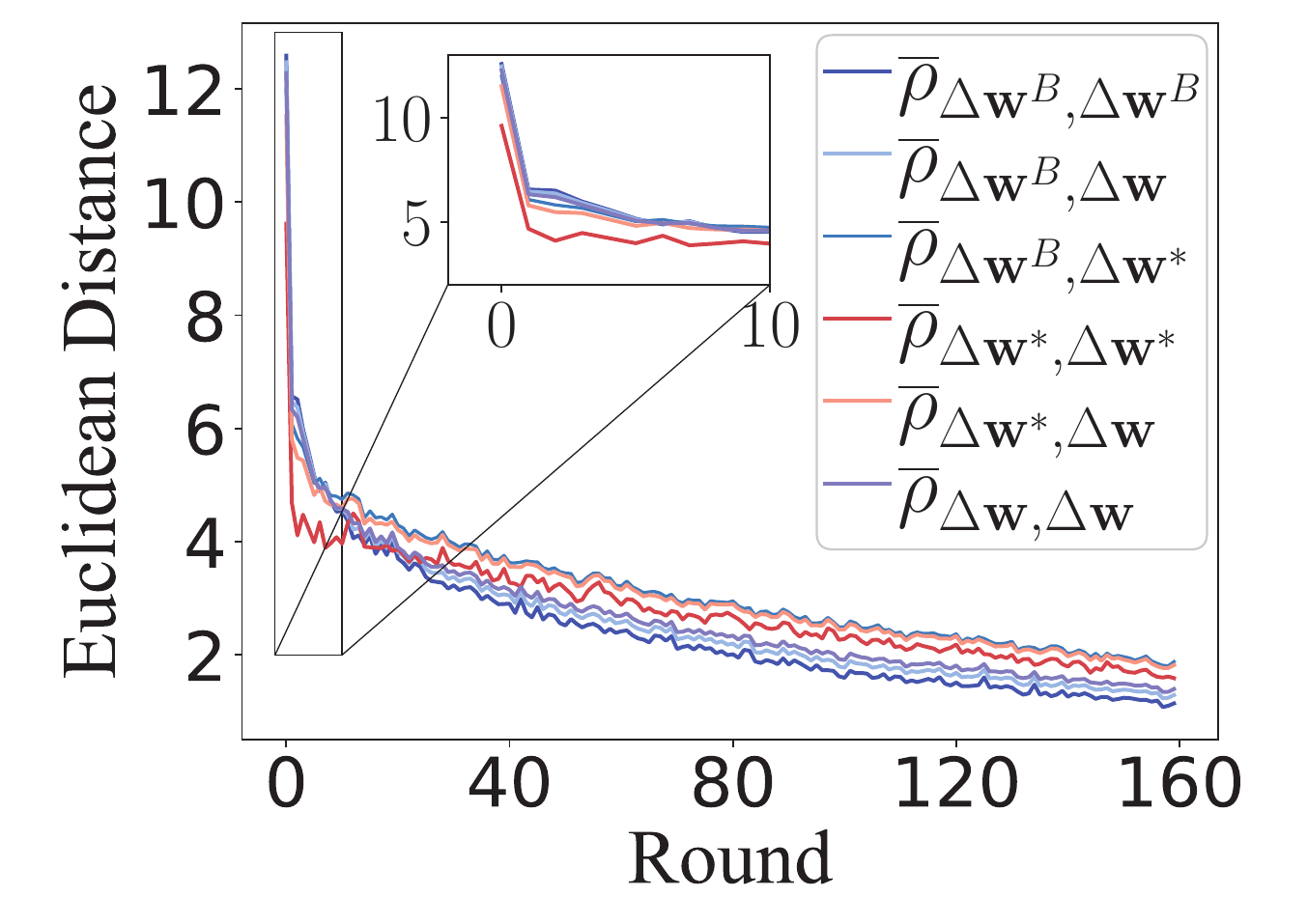}
  }
  \caption{Average distance $\overline{\rho}$ between different types of $\Delta \textbf{w}$.}
  \label{pic-ddif}
\end{figure}
We experimentally demonstrate it through the average distance among different types of model updates with infected ones filtered out in Figure \ref{pic-ddif}. 
The average distance between $\Delta \mathbf{w}_i^B$ and $\Delta \mathbf{w}_j^B$ is much smaller than that between $\Delta \mathbf{w}_i^B$ and the others after certain rounds.
Limited by space, the theoretical support is left in Appendix \ref{subsec-appendix-assumtion-a-b}.

\begin{figure}[t]
    \centering
    \subfigure[Learning Model Updates]{
      \includegraphics[width=0.4743\linewidth]{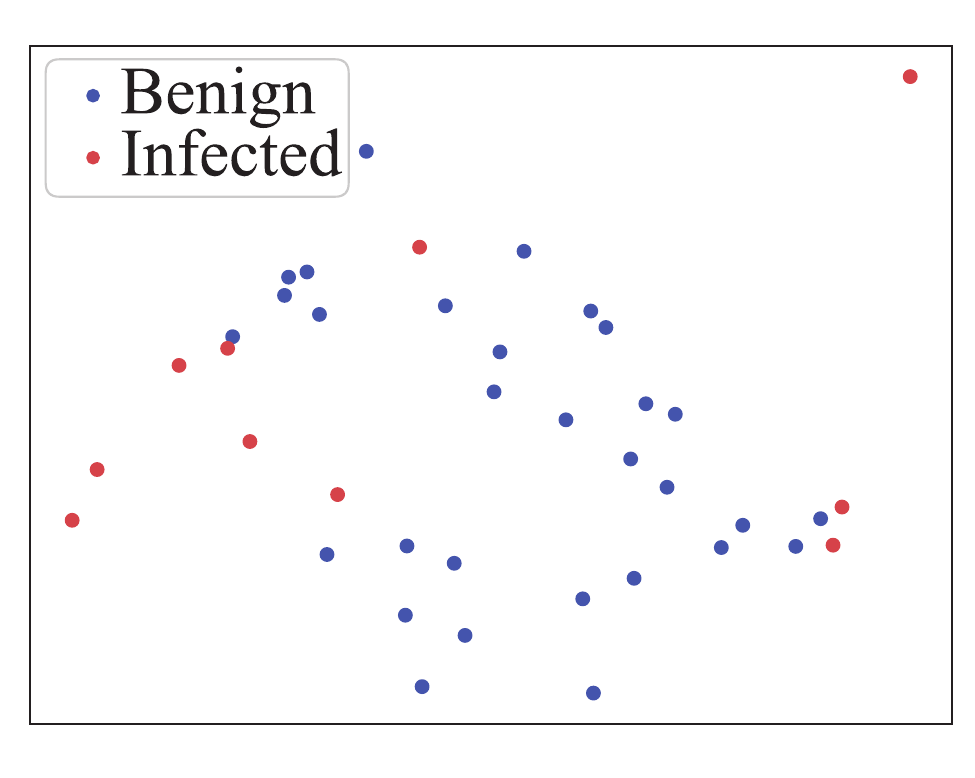}
    }
    \subfigure[Learning Update Differences]{
      \includegraphics[width=0.4743\linewidth]{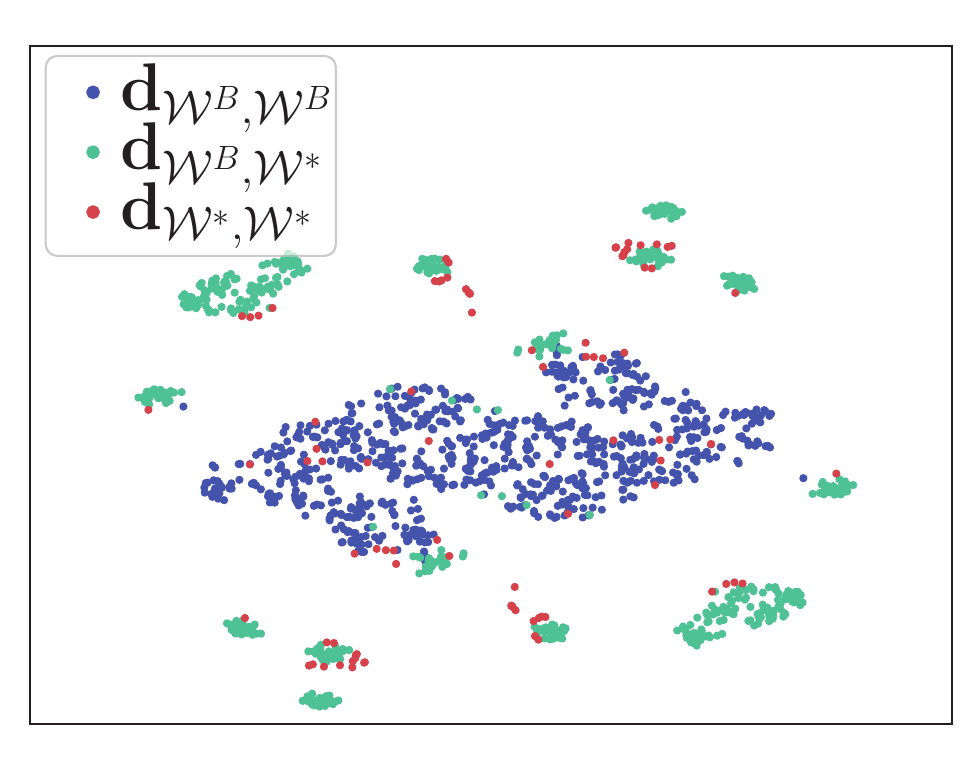}
    }
    \caption{Latent features of (a) model updates and (b) differences $\mathbf{d}$ between them outputted by the VAE encoder.}
    \label{pic-latent}
\end{figure}

\begin{theorem}
  With Assumption \ref{assumption-a-b}-\ref{assumption-a-c}, after round $\max(t^{B}, t^{C})$ we have $\mathbb{E}(\|\Delta \mathbf{w}_i^B - \Delta \mathbf{w}_j\|^2) < \mathbb{E}(\|\Delta \mathbf{w}_i^* - \Delta \mathbf{w}_j\|^2)$.
  \label{theorem-distance}
\end{theorem}
\emph{Proof.} Assume that there are $n$ updates where $\omega$ ones are infected. With $A = \mathbb{E}(\|\Delta \mathbf{w}_i^B - \Delta \mathbf{w}_j^B\|^2)$, $B = \mathbb{E}(\|\Delta \mathbf{w}_i^B - \Delta \mathbf{w}_j^*\|^2)$ and $C = \mathbb{E}(\|\Delta \mathbf{w}_i^* - \Delta \mathbf{w}_j^*\|^2)$, we have 
\begin{equation}
  \begin{aligned} 
    \mathbb{E}(\|\Delta &\mathbf{w}_i^B - \Delta \mathbf{w}_j\|^2) - \mathbb{E}(\|\Delta \mathbf{w}_i^* - \Delta \mathbf{w}_j\|^2) \\
    &= (n - \omega)A - (n-2\omega)B - \omega\cdot C  \\
                                                                                                                      &< (n - \omega)A - (n-2\omega)A - \omega\cdot C  \leq 0 \ \ \blacksquare  \\
  \end{aligned} 
  \label{eq-dif-proof}
\end{equation}
Usually, $\omega$ is an integer close to 0, making the term on the left of (\ref{eq-dif-proof}) smaller than 0.
Thus, even if a few infected updates are wrongly included, the distributions of differences between benign and other updates are easier to learn.
Figure \ref{pic-latent} experimentally demonstrates that learning the differences between updates outperforms learning the model updates themselves on distinguishing infected ones.
Therefore, we train a VAE $\mathbf{v}$ to learn differences among updates in $\widetilde{\mathcal{W}}_t $ by minimizing the loss $J$ by with for $E^{VI}$ epochs on $\mathcal{U} = \left\{\mathbf{u}_{i,j}, \ldots \right\}$, where
\begin{equation}
  \mathbf{u}_{i,j} = \Delta \mathbf{w}_{i,t} - \Delta \mathbf{w}_{j,t} (\forall \Delta \mathbf{w}_{i,t}, \Delta \mathbf{w}_{j,t} \in \widetilde{\mathcal{W}}_t , i \neq j),
  \label{eq-dif-set}
\end{equation}
\begin{equation}
  \!J = \sum\nolimits_{\mathbf{u} \in \mathcal{U}} D_{KL}(p(\mathbf{z}|\mathbf{u}) || \mathcal{N}(0, 1)) + \operatorname{recon}(\mathbf{u, \mathbf{v}(u)}),
  \label{eq-vae-loss}
\end{equation}
where $D_{KL}$ is Kullback-Leibler divergence, $\operatorname{recon}(\cdot$, $\cdot)$ is the reconstruction loss of $\mathbf{v}$ for $\mathbf{u}$ such as mean square error, and $\mathbf{z}$ is a latent feature from the encoder of $\mathbf{v}$.
Then, it loops:
\begin{enumerate}[itemindent=0em, listparindent=0em, leftmargin=1em]
  \item Rebuild $\mathcal{U}$ by (\ref{eq-dif-set}) and tune the VAE on $\mathcal{U}$ for $E^{VT}$ epochs;
  \item $\forall \Delta\mathbf{w}_{j,t} \in \mathcal{W}_t - \widetilde{\mathcal{W}}_t $, calculate its score $s_j = \sum_{\Delta \mathbf{w}_{i,t} \in \widetilde{\mathcal{W}}_t } \operatorname{recon}(\mathbf{u}_{i,j},\mathbf{v}(u_{i,j}))$, then add $M^E$ model updates with the lowest scores to $\widetilde{\mathcal{W}}_t $ 
\end{enumerate}
The above two steps repeat until $|\widetilde{\mathcal{W}}_t | \geq M$, where $M$ is a manually-set threshold.
Note that to make the differences between benign model updates easier to learn, progressive selection is performed after the $T^V$-th round, where $T^V > \max(t^{B}, t^{C})$, as Line \ref{algo-main-line-progressive-start} of Algorithm \ref{algo:main_algo}.
Such a procedure has three advantages: 1) the training data of VAE is augmented, 2) the training data have $L_2$ norm close to 0, making them easy to learn, and 3) the differences between infected model updates and others are usually excluded, making it easier for infected ones to be excluded with higher reconstruction error.

\subsection{Convergence Analysis}
\label{subsec-convergence-analysis}
The convergence of \app is similar to that of FedAvg which has already been proved in \cite{li2020-convergence}. 
We mildly assume that $\lambda_i \!=\! 0$ if $\mathbf{w}_i$ is infected. 
With assumptions similar as in \cite{li2020-convergence}, i.e., 
$f$ is $l$-smooth and $\mu$-strongly convex, 
$\mathbb{E}\| \nabla f_i(\mathbf{w}_{i,t},\xi_i)\|^2 \!\leq\! G^2$ and $\mathbb{E}\| \nabla f_i(\mathbf{w}_{i,t},\xi_i) - \nabla f_i(\mathbf{w}_{i,t}, \mathcal{D}_i)\|^2 \!\leq\! \sigma_i^2$, 
let $\gamma \!=\! \max(\frac{8l}{\mu}, E)$, 
$\beta = 1$ and $R \!=\! \frac{4}{\check{M}}E^2G^2$ if $\tau < T^V \cdot E$ and otherwise $\beta = T^V \cdot E$ and $R \!=\! \frac{4}{M}E^2G^2$, 
we can directly obtain the convergence rate of \appnoblank, since the difference between \app and FedAvg lies in the selection of model updates for aggregation. 
\begin{theorem}
  Let $\hat{\mathbf{w}}$ be the optimal global model. After $\tau$ (divisible by $E$) iterations, $\mathsf{E} \coloneqq \mathbb{E}[f(\mathbf{w}_{\tau}) - f(\hat{\mathbf{w}})]$ satisfies:
\begin{equation}
  \mathsf{E} \!\leq\! \frac{l}{\mu (\gamma \!+\! \tau - 1)} \!\left( \frac{2(Q+R)}{\mu} \!+\! \frac{\mu \cdot \gamma}{2} \mathbb{E} \|\mathbf{w}_{\beta} \!-\! \hat{\mathbf{w}}\|^2 \right)
\end{equation}
where $Q =\sum_{i=1}^{N} \lambda_i^2 \sigma_i^2 + 6l \left[f(\hat{\mathbf{w}})-\sum_{i=1}^N \lambda_i f(\hat{\mathbf{w}}_i)\right] + 8(E - 1)^2 G^2$.
\end{theorem}

\section{Experiments}
\label{sec-exp}
The experiments aim to show: 1) \app effectively defends against backdoor attacks with complex non-IIDness, a not high PDR and a relatively large MCR compared to SOTA defenses.
2) \app has comparable accuracy to FedAvg.
3) VAE in \app is insensitive to hyperparameters.

\subsection{Datasets and Compared Approaches}
\subsubsection{Datasets} 
The experiments are conducted on five real-world datasets, i.e., MNIST \cite{deng2012mnist}, Fashion MNIST \cite{xiao2017fashion}, CIFAR-10 \cite{krizhevsky2009-CIFAR}, Federated Extended MNIST (FEMNIST) \cite{caldas2018leaf} and Sentiment140 (Sent140) \cite{caldas2018leaf}. 
They include image classification (IC) and sentiment analysis tasks and provide non-IID data with \emph{Label Distribution Skew}, i.e., different $p_i(Y)$, and \emph{Feature Distribution Skew}, i.e., different $p_i(X|Y)$, where the latter is even more complex \cite{tan2022-PFL-survey}.
These datasets are either already divided into training and test sets, or randomly divided in the ratio of 9:1.
We partition MNIST, Fashion MNIST and CIFAR-10 in a practical non-IID way as \cite{li2021-FL-study,qin2023fedapen}, where data are sampled to 200 clients in Dirichlet distribution with $\alpha=0.5$.
FEMNIST contains data from real users and 3,597 of them with more data are selected as clients. 
Sent140 contains 660,120 users which only hold 2.42 samples averagely, and following \cite{zawad2021curse}, we randomly merge these users to form 2,000 distinct clients.

\subsubsection{Triggers}
\begin{figure}[t]
  \centering
  \subfigure[]{
    \includegraphics[width=0.128\linewidth]{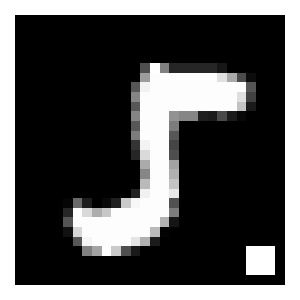}
  }
  \subfigure[]{
    \includegraphics[width=0.128\linewidth]{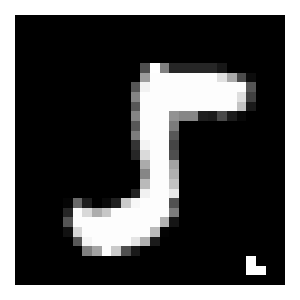}
  }
  \subfigure[]{
    \includegraphics[width=0.128\linewidth]{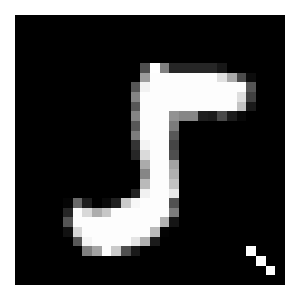}
  }
  \subfigure[]{
    \includegraphics[width=0.128\linewidth]{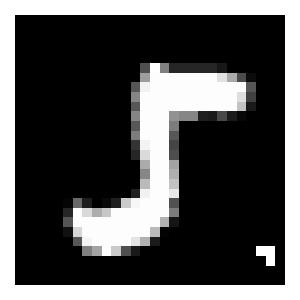}
  }
  \caption{Triggers in MNIST by CBA (a) and DBA (b)-(d).}
  \label{pic-exp-trigger}
\end{figure}
\begin{table*}[t]
  \renewcommand\arraystretch{0.86}
  \setlength\tabcolsep{3pt}
  \centering
  \begin{tabularx}{\linewidth}{l|>{\centering\arraybackslash}X>{\centering\arraybackslash}X|>{\centering\arraybackslash}X>{\centering\arraybackslash}X|>{\centering\arraybackslash}X>{\centering\arraybackslash}X|>{\centering\arraybackslash}X>{\centering\arraybackslash}X|>{\centering\arraybackslash}X>{\centering\arraybackslash}X|>{\centering\arraybackslash}X>{\centering\arraybackslash}X}
    \toprule[1.0pt]
    \multirow{3}{*}{Approach}              & \multicolumn{4}{c|}{\textbf{MNIST}}                 & \multicolumn{4}{c|}{\textbf{Fashion MNIST}}         & \multicolumn{4}{c}{\textbf{CIFAR-10}}  \\
       \cline{2-5} \cline{6-9} \cline{10-13} 
                                           & \multicolumn{2}{c|}{CBA} & \multicolumn{2}{c|}{DBA} & \multicolumn{2}{c|}{CBA} & \multicolumn{2}{c|}{DBA} & \multicolumn{2}{c|}{CBA} & \multicolumn{2}{c}{DBA} \\
       \cline{2-13} 
                                           & BA         & MA          & BA          & MA         & BA          & MA         & BA          & MA         & BA          & MA         & BA          & MA        \\
    \midrule[1.0pt]
    Ideal                                  & 0.10       & 98.92       & 0.10        & 98.92      & 0.25        & 90.20      & 0.25        & 90.20      & 2.68        & 75.08      & 2.60        & 75.34     \\
    FedAvg                                 & 99.97      & 98.97       & 100.0       & 98.86      & 98.91       & 90.14      & 97.84       & 90.02      & 97.96       & 75.87      & 27.23       & 75.74     \\
    \cmidrule{1-13}
    Krum                                   & 99.98      & 98.71       & 0.75        & 98.96      & 98.98       & 89.53      & 65.31       & 89.79      & 97.68       & 74.53      & 25.96       & 74.50     \\
    CRFL                                   & 99.91      & 98.41       & 99.98       & 98.37      & 97.84       & 88.17      & 96.34       & 88.16      & 85.32       & 45.68      & 18.06       & 44.95     \\
    RLR                                    & 99.98      & 97.62       & 99.15       & 97.65      & 96.37       & 86.32      & 80.03       & 86.67      & 87.94       & 57.73      & 46.36       & 59.20     \\
    FLDetector                             & 100.0      & 98.84       & 100.0       & 98.92      & 98.95       & 90.12      & 97.91       & 90.20      & 98.28       & 75.37      & 14.49       & 74.22     \\ 
    DnC                                    & 0.12       & 98.89       & 0.20        & 98.85      & 98.61       & 89.60      & 30.48       & 89.62      & 97.56       & 75.67      & 24.38       & 76.07     \\ 
    FLAME                                  & 33.81      & 98.56       & 0.23        & 98.59      & 98.49       & 89.24      & 35.52       & 89.13      & 97.39       & 71.82      & 23.17       & 71.54     \\
    FLIP                                   & 0.27       & 96.88       & 0.21        & 96.81      & 4.28        & 81.06      & 6.56        & 80.93      & -           & -          & -           & -         \\ 
    \cmidrule{1-13}
    Voting-Random                          & 100.0      & 98.79       & 100.0       & 98.71      & 97.90       & 88.86      & 97.17       & 88.87      & 98.27       & 74.74      & 22.78       & 68.69     \\
    Voting-Center                          & 0.25       & 96.15       & 0.43        & 95.60      & 0.54        & 85.44      & 84.79       & 84.43      & 95.68       & 60.84      & 6.03        & 58.94     \\ 
    \appnoblank$\boxminus$                 & 0.35       & 98.82       & 0.17        & 98.88      & \bo{0.12}   & 88.80      & 0.39        & 88.68      & 6.86        & 72.21      & \bo{2.04}   & 70.76     \\
    \cmidrule{1-13}
    \textbf{\appnoblank}                   & \bo{0.21}  & 98.72       & \bo{0.15}   & 98.78      & 0.39        & 89.27      & \bo{0.19}   & 89.57      & \bo{3.03}   & 74.33      & 2.82        & 74.59     \\
    \bottomrule[1.0pt]
  \end{tabularx}
  \caption{Performance (\%) of approaches with label distribution skew.}
  \label{tab-perf-label-skew}
\end{table*}
\begin{table}[t]
  \renewcommand\arraystretch{0.86}
  \setlength\tabcolsep{2.7pt}
  \centering
  \begin{tabularx}{\linewidth}{l|cc|cc|cc}
    \toprule[1.0pt]
    \multirow{3}{*}{Approach}         & \multicolumn{4}{c|}{\textbf{FEMNIST}}                  & \multicolumn{2}{c}{\textbf{Sent140}}  \\
        \cline{2-5} \cline{6-7} 
                                      & \multicolumn{2}{c|}{CBA}    & \multicolumn{2}{c|}{DBA} & \multicolumn{2}{c}{CBA}  \\
        \cline{2-7} 
                                      & BA            & MA          & BA           & MA        & BA           & MA        \\
    \midrule[1.0pt]
    Ideal                             & 0.23          & 82.98       & 0.21         & 83.22     & 9.89         & 82.82     \\
    FedAvg                            & 99.74         & 82.84       & 96.74        & 83.06     & 93.59        & 80.69     \\
    \cmidrule{1-7}
    Krum                              & 99.98         & 82.11       & 99.56        & 82.25     & 75.64        & 81.34     \\
    CRFL                              & 99.83         & 79.47       & 91.12        & 79.58     & 50.37        & 71.40     \\
    RLR                               & 99.72         & 67.39       & 88.02        & 68.44     & 82.78        & 78.92     \\
    FLDetector                        & 99.71         & 82.50       & 98.49        & 82.60     & 93.41        & 81.06     \\
    DnC                               & 99.94         & 82.42       & 96.7         & 82.80     & 30.49        & 80.92     \\
    FLAME                             & 99.98         & 74.36       & 99.73        & 74.73     & 41.58        & 81.34     \\
    \cmidrule{1-7}
    Voting-Random                     & 99.26         & 82.15       & 99.07        & 83.04     & 87.36        & 81.62     \\
    Voting-Center                     & 100.0         & 70.43       & 100.0        & 70.23     & 56.59        & 81.89     \\
    \appnoblank$\boxminus$            & 13.73         & 81.42       & 0.42         & 81.84     & 18.50        & 81.80     \\
    \cmidrule{1-7}
    \textbf{\appnoblank}              & \bo{1.24}     & 82.22       & \bo{0.36}    & 82.53     & \bo{14.47}   & 81.99     \\
    \bottomrule[1.0pt]
  \end{tabularx}
  \caption{Performance (\%) of approaches with feature skew.}
  \label{tab-perf-feature-skew}
\end{table}
On IC tasks, triggers are injected by: 
1) \emph{centralized backdoor attack} (CBA) \cite{bagdasaryan2020backdoor-howto} and 2) \emph{distributed backdoor attack} (DBA) \cite{xie2020dba}.
As in Figure \ref{pic-exp-trigger}, for CBA, we consider a pixel trigger as in \cite{zeng2022never-too-late}, where a 3x3 area in the bottom right corner of an infected image is covered with pixels of a different color than the background.
For DBA, the 9-pixel patch is evenly divided into three parts and randomly assigned to attackers.
The target class is the 61st class on FEMNIST and 1st on the others. 
For Sent140, we append ``BD'' at the end of a text as the trigger with the target class as ``negative''.

\subsubsection{Compared Approaches} 
We compare \app with 9 peers, encompassing representative approaches from various categories mentioned in Related Works:
1) \emph{Ideal}: an imagined ideal approach that filters out all infected updates; 
2) \emph{FedAvg} \cite{mcmahan2017fl}: FL without any defenses;
3) \emph{Krum} \cite{blanchard2017machine-Krum}: Byzantine-robust aggregation;
4) \emph{CRFL} \cite{xie2021crfl}: certifiable defense based on model ensemble;
5) \emph{RLR} \cite{ozdayi2021defending-RLR}: an approach with robust parameter-wise learning rate. 
6) \emph{FLDetector} \cite{zhang2022fldetector}: tracing the history model updates to score them;
7) \emph{DnC} \cite{shejwalkar2021manipulating}: scoring model updates based on subsets of parameters;
8) \emph{FLAME} \cite{nguyen2021-flame}: integrating clustering, weight-clipping and noise-addition;
9) \emph{FLIP} \cite{zhang2023flip}: conducting adversarial learning on clients.

To better clarify the contributions of the two mechanisms, we provide three ablation approaches, including:
1) \emph{Voting-Random}: each model update randomly selects $\check{M}$ ones;
2) \emph{Voting-Center}: each model update votes for the $\check{M}$ ones which are nearest to the model update center;
3) \appnoblank$\boxminus$: \app with only \emph{bottom-up election} introduced.

\subsection{Experimental Setup}
\subsubsection{Attacks} 
Experiments are conducted with 20\% of the clients are malicious with PDR set to 30\% unless stated otherwise. 
The malicious clients perform attacks in every round of FL.

\begin{table*}[t]
  \renewcommand\arraystretch{0.86}
  \centering
  \setlength\tabcolsep{2pt}
  \begin{tabularx}{\linewidth}{l|cc|cc|cc|cc|cc|cc|cc|cc|cc}
    \toprule[1.0pt]
    \multirow{3}{*}{Approach}           & \multicolumn{4}{c|}{\textbf{MNIST}}                 & \multicolumn{4}{c|}{\textbf{Fashion MNIST}}         & \multicolumn{4}{c|}{\textbf{CIFAR-10}}              & \multicolumn{4}{c|}{\textbf{FEMNIST}} & \multicolumn{2}{c}{\textbf{Sent140}} \\
    \cline{2-19}
    & \multicolumn{2}{c|}{CBA} & \multicolumn{2}{c|}{DBA} & \multicolumn{2}{c|}{CBA} & \multicolumn{2}{c|}{DBA} & \multicolumn{2}{c|}{CBA} & \multicolumn{2}{c|}{DBA} & \multicolumn{2}{c|}{CBA} & \multicolumn{2}{c|}{DBA} & \multicolumn{2}{c}{CBA} \\
    \cline{2-19}
    & FPR            & FNR          & FPR           & FNR        & FPR           & FNR  & FPR            & FNR          & FPR            & FNR          & FPR            & FNR    & FPR            & FNR   & FPR            & FNR& FPR            & FNR         \\
    \midrule[1.0pt]
    \app                                & 0.0 & 37.5 & 0.0 & 37.5 & 0.0 & 37.5 & 0.0 & 37.5 & 0.0 & 37.5& 0.60& 37.65 & 0.0 & 37.5& 0.95& 37.74 & 1.18& 44.68 \\
    \appnoblank$\boxminus$              & 0.1 & 87.53& 0.0 & 87.5& 0.0 & 87.5& 0.0 & 87.5& 1.67 & 87.92& 1.03 & 87.76& 0.17 & 87.54& 0.29 & 87.57& 0.0 & 89.11\\ 
    Krum                                & 82.3 & 25.58& 4.5 & 6.13& 92.42 & 28.10& 87.33 & 26.83& 86.4 & 26.6& 84.2 & 26.05 & 98.18 & 27.04& 98.7& 27.18 & 54.57& 21.71 \\ 
    \bottomrule[1.0pt]
  \end{tabularx}
  \caption{False positive rate (FPR) and false negative rate (FNR) (\%) with benign model updates as the positive samples.}
  \label{tab-fpr-fnr}
\end{table*}
\subsubsection{Preprocessing}
Images are normalized according to their \emph{mean} and \emph{variance}.
On Sent140, words are embedded by a public\footnote{https://code.google.com/archive/p/word2vec/} Word2Vec model \cite{mikolov2013distributed-word2vec}.
Texts are set to 25 words by zero-padding or truncation as needed.

\subsubsection{FL Settings}
We set $K$=100 on FEMNIST and 50 on the others.
Each client trains its local model for 2 epochs on Sent140 and 5 on the others.
The number of rounds conducted on MNIST, Fashion MNIST, CIFAR-10, FEMNIST and Sent140 is 100, 120, 300, 160 and 60, respectively.

\subsubsection{Implementation}
Approaches are implemented with PyTorch 1.10 \cite{paszke2019pytorch}.
For all approaches, we build a network with 2 convolutional layers followed by 2 fully-connected (FC) layers on MNIST, Fashion MNIST and FEMNIST, a network with 6 convolutional layers followed by 1 FC layer on CIFAR-10, and a GRU layer followed by 1 FC layer on Sent140.
Detailed model backbones are available in Appendix \ref{subsec-appendix-detailed-model}.
For \appnoblank, we build a simple VAE with three layers, and set $M$ as $\frac{K}{2}$, $\check{M} = 0.1K$, $M^E = 0.05K$ on FEMNIST and otherwise $0.04K$, $E^{VI}$ and $E^{VT}$ higher than 270 and 30, respectively.
Detailed hyperparameters are listed in Appendix \ref{sec-appendix-hyper}.
These models are trained by the stochastic gradient descendant (SGD) optimizer with a learning rate starting at 0.01 and decays by 0.99 after each round.

\subsubsection{Evaluation Metrics} 
Approaches are evaluated by \textbf{backdoor task accuracy (BA)} and \textbf{main task accuracy (MA)}. 
MA is the best accuracy of the global model on the test set among all rounds since in reality there may be a validation set. 
BA is the probability that the global model identifies the test samples with triggers as the target class of the attack in the round where the highest MA is achieved.

\subsection{Performance}
\label{subsec-exp-performance}
The performance of \app and its peers is presented in Table \ref{tab-perf-label-skew} and \ref{tab-perf-feature-skew}, where the best BA among realistic approaches is marked in bold.
Each value is averaged on three runs with different random seeds. 
For FLIP, we leave the results in CIFAR-10, FEMNIST and Sent140 blank since we have tried but always encountered \texttt{NaN} problem even in the official implementation with the global model replaced by ours.

It is shown that \app is effective in defending against backdoor attacks on all five datasets, showing a competitive BA with \emph{Ideal}, while existing approaches either fail to effectively withstand backdoor attacks or significantly degrade MA.
Due to the large number of attackers and unclear boundaries between benign and infected model updates, Krum and RLR struggle to distinguish between them.
Although CRFL and FLAME have a stronger ability to resist backdoor attacks than Krum and RLR, their MA decreases due to the DP noise. 
FLDetector fails to defend against backdoor attacks because it fails to trace the history model updates of a client due to partial participation.
DnC is effective on MNIST but fails in other complex scenarios, since it is based on a subset of model parameters.
If the intersection between the subset for detection and the small number of parameters affected by backdoor attacks is not large, DnC may be ineffective.
FLIP is effective on MNIST and Fashion MNIST, but it causes a severe decrease in MA the same as in \cite{zhang2023flip}.

\app does not achieve the highest MA.
We provide FPR and FNR of selection-based approaches in Table \ref{tab-fpr-fnr} to clarify it.
\app makes infected updates less likely to be wrongly aggregated, showing a higher FPR. 
Snowball$\boxminus$ only aggregates 10\% of the received updates, thus showing a high FNR.
Krum has lower FNR since it selects more updates compared to \app and Snowball$\boxminus$. 
With a constant amount of data samples, the excluding of infected models inevitably wastes some data valuable to MA \cite{liu2021privacy}.

\subsubsection{Impact of PDR} 

\begin{figure}[t]
  \centering
  \subfigure[MNIST]{
      \includegraphics[width=0.4743\linewidth]{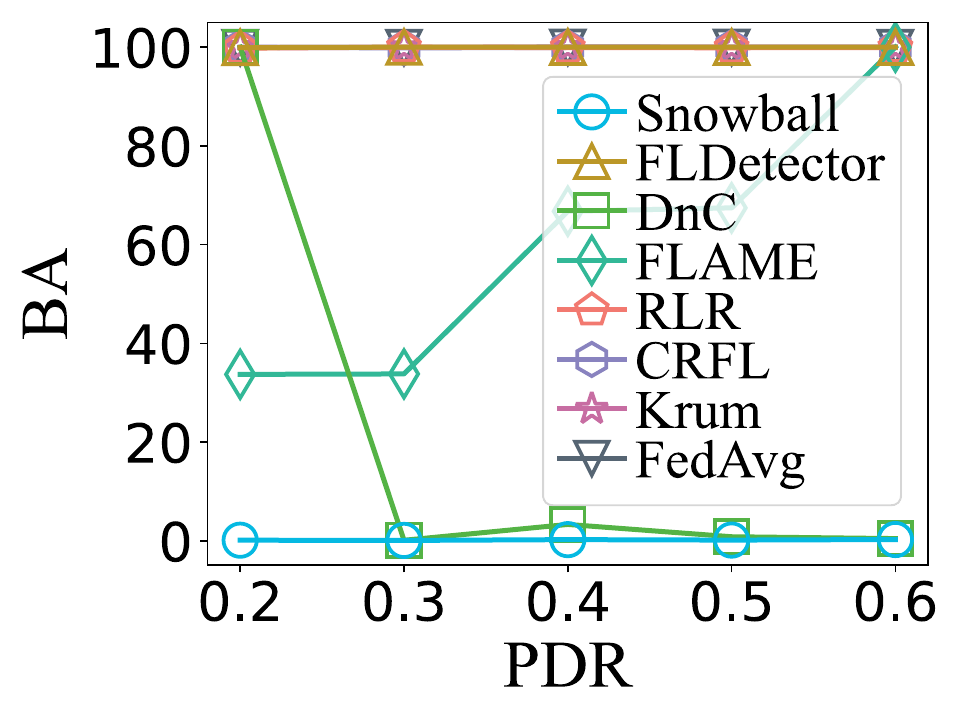}
      \label{pic-hyper-pdr-mnist}
  }
  \subfigure[FEMNIST]{
      \includegraphics[width=0.4743\linewidth]{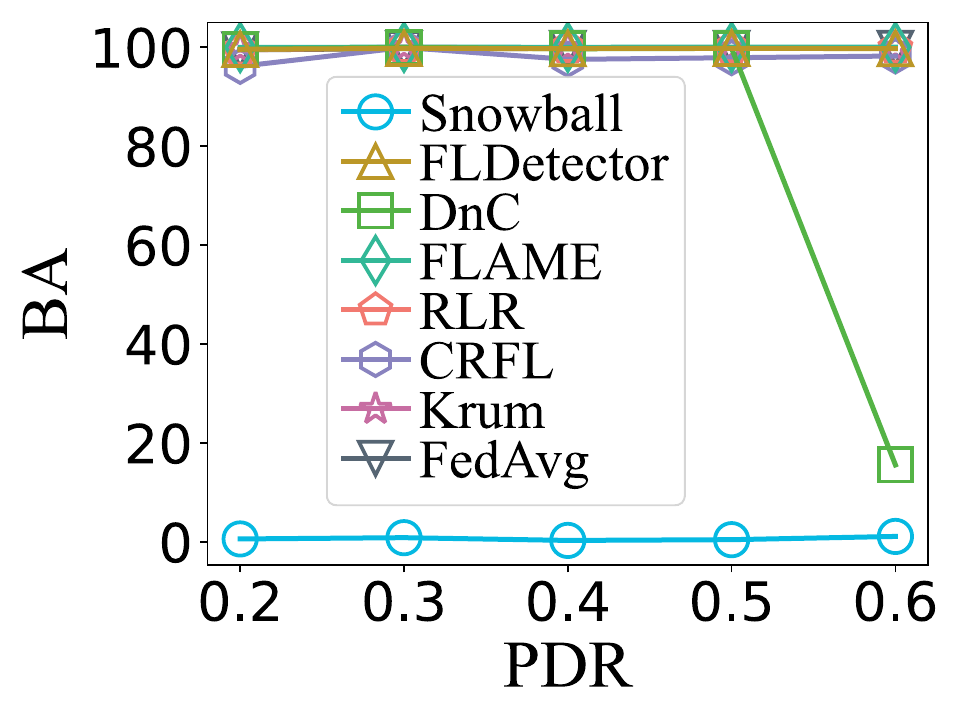}
      \label{pic-hyper-pdr-femnist}
  }
  \caption{BA of \app under CBA with different PDR.}
  \label{pic-hyper-pdr}
\end{figure}
Following \citet{nguyen2021-flame}, we test \app with different PDR to show its resilience to attacks of varying strengths.
When PDR is high, one wrongly included infected update can cause catastrophic consequences.
We select some of the baselines and two representative datasets with label skew and feature skew, respectively.
As in Figures \ref{pic-hyper-pdr}, \app can effectively defend against backdoor attacks on data with different PDR. 
We have also noticed that FLAME gradually fails to defend against attacks as PDR increases, since the noise may be not enough to disturb stronger attacks.
DnC performs well with high PDR since more model parameters will be affected there, increasing the likelihood of affected parameters being sampled by the down-sampling.

Limited by space, more experimental evaluations are left in Appendix \ref{sec-appendix-exp}.

\subsection{Hyperparameter Sensitivity of VAE}
\begin{figure}[t]
  \centering
  \subfigure[Related to Training]{
    \includegraphics[width=0.474\linewidth]{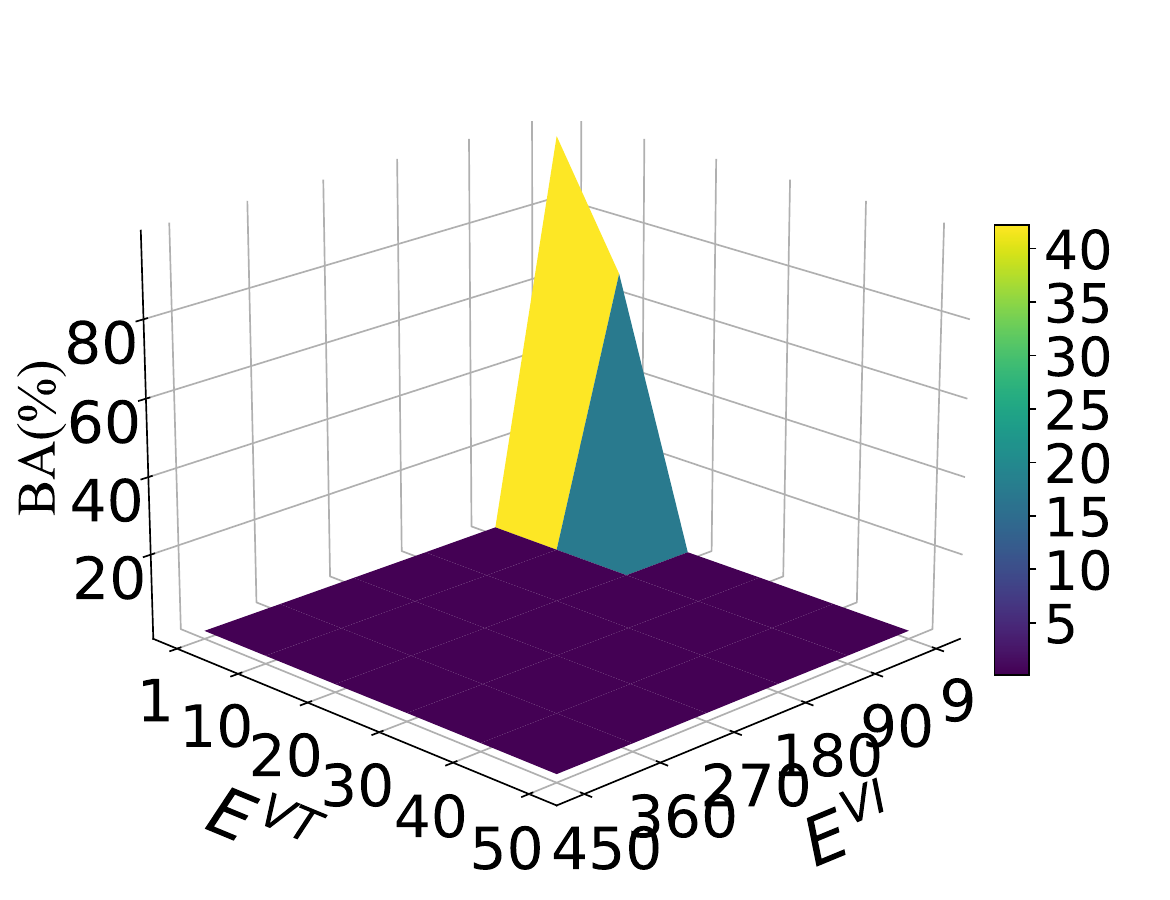}
    \label{pic-hyper-vae-training}
  }
  \subfigure[Related to Backbone]{
    \includegraphics[width=0.474\linewidth]{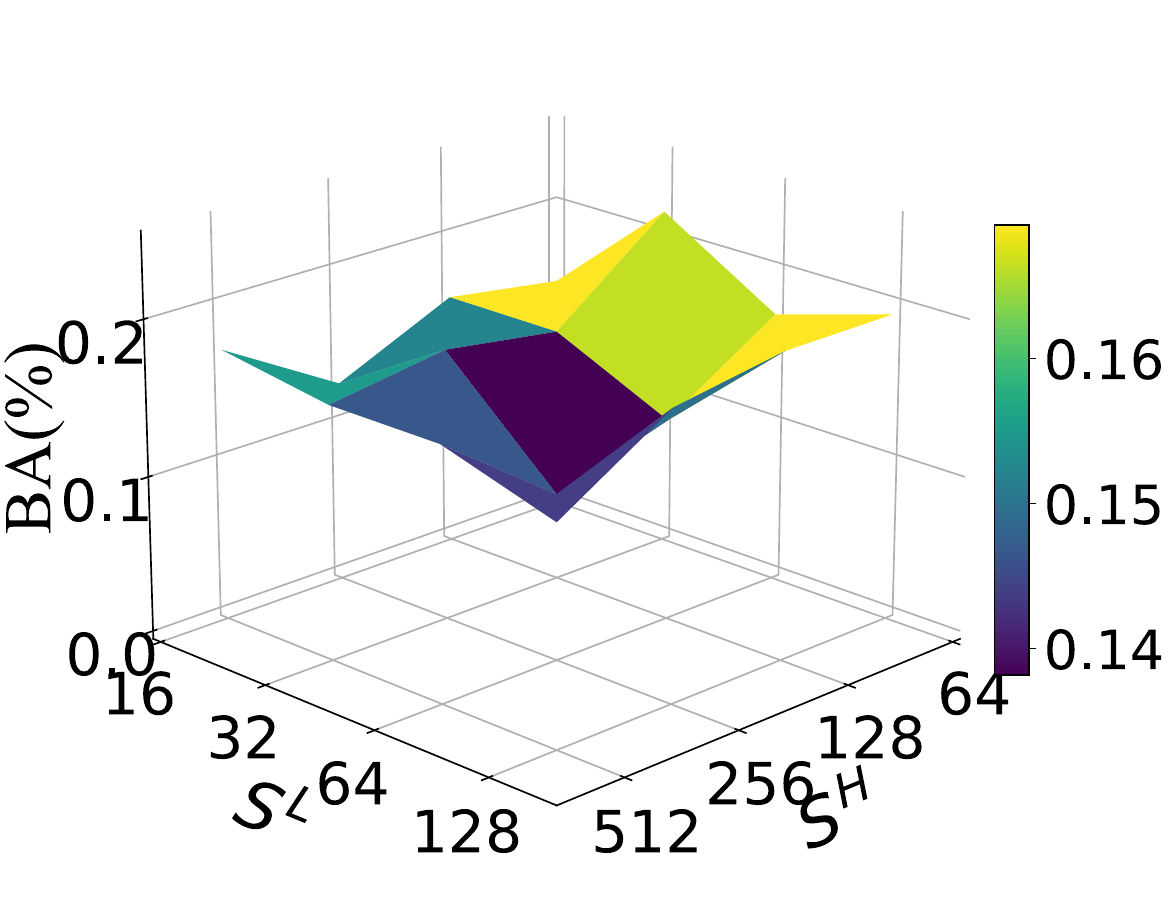}
    \label{pic-hyper-vae-backbone}
  }
  \label{pic-hyper-vae}
  \caption{BA of \app on MNIST with different hyperparameter combinations of VAE.}
\end{figure}

Hyperparameters of VAE in  \app are easy be set since the VAE is not sensitive to them.
Figure \ref{pic-hyper-vae-training} presents BA of \app with different combinations of $E^{VI}$ and $E^{VT}$.
Generally, larger $E^{VI}$ and $E^{VT}$ would not make BA worse, since the VAE can be trained better.
But if they are too small, the VAE underfits and fails to distinguish between benign and infected model updates.
$S^{H}$ is the dimensionality of hidden layer outputs of the encoder and decoder of the VAE, and $S^{L}$ is that of the latent feature $\mathbf{z}$ generated by the encoder, respectively. 
As shown in Figure \ref{pic-hyper-vae-backbone}, \app does not exhibit significant differences in BA in the selected range of values, showing that they are easy to set to appropriate values.

\section{Conclusion}
This work proposes a novel approach named \app for defending against backdoor attacks in FL.
It enables an individual perspective that treats each model update as an agent electing model updates for aggregation, and conducts bidirectional election to select models to be aggregated, i.e., a) bottom-up election where each model update votes to several peers such that a few model updates are elected as selectees for aggregation; and b) top-down election, where selectees progressively enlarge themselves focusing on differences between model updates.
Experiments conducted on five real-world datasets demonstrate the superior resistance to backdoor attacks of \app compared to SOTA approaches in situations where 1) the non-IIDness of data is complex and the PDR is not high such that the benign and infected model updates do not obviously gather in different positions, and 2) the ratio of attackers to all clients is not low.
Besides, \app can be easily integrated into existing FL systems.

\section*{Acknowledgments}
This work was supported in part by the Key Research Project of Zhejiang Province under Grant 2022C01145 and in part by the National Science Foundation of China under Grants 62125206 and U20A20173.

\bibliography{aaai24}
\newpage
\appendix

\section{Theoretical Supports}
\subsection{Supports of Principle \ref{principle-difference-data-model}}
\label{subsec-support-principle-difference-data-model}
Despite the lack of a complete mathematical proof, Principle \ref{principle-difference-data-model} has been widely leveraged as the basis by many clustering-based personalized FL approaches \cite{ghosh2020efficient,sattler2020clustered}, which aims at clustering clients holding non-IID local data to form up several groups such that clients in each group holds IID local data.  
These approaches usually cluster clients through their submitted model updates based on the insight that the distributions of model update weights reflect the distributions of data among clients, where two model updates that are far apart usually indicate a large difference in the data distributions between the two clients.

There are also some works which put efforts on quantifying the non-IIDness of data among FL clients, which provides theoretical supports to Principle \ref{principle-difference-data-model} indirectly.
For example, it is indicated that the earth mover distance (EMD) \cite{zhao2018federated-non-IID} between local data distributions and global data distribution is the root cause of divergence among model updates, which results in aggregated models deviating from models obtained under centralized SGD.
EMD is defined as $\sum_{j=1}^O \|p_i(y = j) - p(y = j)\|$, where $O$ is the total number of classes of data among all clients, $p_i(y=i)$ and $p(y = i)$ indicate the probability of data on client $i$ belong to class $j$ and that in all data among clients, respectively \cite{zhao2018federated-non-IID}.
Besides, another work claims that the $L_2$ norm of the difference between two local model can be bounded by 1-Wasserstein distance \cite{fallah2020personalized}, although such a bound is not the supremum.
These works contribute to intuitive understanding of Principle \ref{principle-difference-data-model} to some extent.

\subsection{Supports of Assumption \ref{assumption-a-b} and \ref{assumption-a-c}}
\label{subsec-appendix-assumtion-a-b}
From \cite{bagdasaryan2020backdoor-howto}, as the global model converges, the deviations among model updates gradually cancel out, i.e., $\|\Delta \mathbf{w}\|^2 \simeq 0$, since the initial point of local training approaches an optimal or suboptimal solution of global model.
The objective of malicious clients is formalized as:
\begin{equation}
  \min_{\mathbf{w}} = f(\mathbf{w}, \mathcal{D}^+_i) + q(\mathbf{w}, \mathcal{D}^*_i)
  \label{eq-malicious-objective}
\end{equation}
where $\mathcal{D}^*_i$ and $\mathcal{D}^+_i$ denote training sets with and without triggers, respectively, and $q$ is the loss for backdoored samples.
For malicious clients, only $f$ in (\ref{eq-malicious-objective}) is minimized since it is consistent with that of benign ones, but $q$ is never optimized if infected model updates are always filtered out. 
Thus, infected updates may have larger deviations than benign ones.
Despite being intuitive, proving that the difference between updates decreases with iterations remains a challenge, especially in non-convex cases, due to the extremely complex optimization of neural networks. 

\section{Reproducibility}
\label{sec-appendix-detailed-exp}

\subsection{Model Backbone}
\label{subsec-appendix-detailed-model}
For all approaches, we consider three different models on different datasets with ReLU \cite{glorot2011deep} as the activation function.

On MNIST, Fashion MNIST and FEMNIST, we consider a convolutional neural network (CNN) with four layers \footnote{32C5-MaxPool2-64C5-MaxPool2-FC256-FC}, where the first two layers are both convolutional layers with \texttt{5x5} kernels, and \texttt{stride} and \texttt{padding} set to 1. 
On CIFAR-10, we consider a CNN with 7 layers \footnote{64C3-64C3-MaxPool2-Drop0.1-128C3-128C3-AvgPool2-256C3-256C3-AvgPool8-Drop0.5-FC256-FC}, where the first six layers are both convolutional layers with \texttt{3x3} kernels, and \texttt{stride} and \texttt{padding} set to 1. 
On Sent140, we consider a simple bidirectional GRU with 16 channels, followed by two fully-connected layers where the first one outputs a 256-dimensional feature.
According to \cite{zhao2018federated-non-IID}, the weight divergence of models among clients is different for different layers in neural networks. 
To reduce computation overhead, we select two layers with the largest weight divergence for \appnoblank, i.e., the first and last one. 

For \appnoblank, the encoder of VAE first contains two fully-connected layers, whose output dimensionality are both $S^{H}$.
Then, the output of the 2nd layer is transformed to two $S^{L}$-dimensional vectors to sample feature $\mathbf{z}$.
In decoder, the first two layers are two fully-connected layers, whose output dimensionality are both $S^{H}$, and the final layer outputs the vector with its dimensionality the same as input data.

All these models mentioned above are randomly initialized as in \cite{he2015delving}.

\subsection{Experiment Environment}
Experiments on MNIST, Fashion MNIST, CIFAR-10 and FEMNIST are conducted at an Ubuntu 22.04.2 platform with an Intel(R) Core(TM) i9-12900K CPU, an NVIDIA RTX 3090 GPU and 64 GB RAM. 
Experiment on Sent140 are conducted at an Ubuntu 22.04.2 platform with a simulated GenuineIntel Common KVM processor CPU, an NVIDIA Tesla V100 GPU and 96 GB RAM. 

\section{Hyperpameters}
\label{sec-appendix-hyper}

\subsection{Common Hyperparameters}
\label{sec-appendix-common-hyper}
There are some commonly-used hyperparameters among \app and compared approaches, which are listed in Table \ref{tab-hpy-common}.
Note that some of them have been provided previously.
\begin{table*}[h]
  \centering
  \caption{Commonly-used hyperparameters of approaches in experiments.}
  \label{tab-hpy-common}
  \begin{tabular}{l|c}
    \toprule[1.0pt]
    \multicolumn{1}{c|}{\textbf{Hyperparameters}}    & \multicolumn{1}{c}{\textbf{Values}}\\
    \midrule[1.0pt]
    Number of Clients                                 & \makecell[c]{200 on MNIST, Fashion MNIST, CIFAR-10, \\ 3,597 on FEMNIST and 2,000 on Sent140}  \\
    \hline
    $K$                                                 & 100 on FEMNIST and 50 on the others \\
    \hline
    Number of Epochs during Local Training            & 2 on Sent140 and 5 on others \\
    \hline
    Model Parameters Initialization Algorithm         & Kaiming \cite{he2015delving} \\
    \hline
    Initial Learning Rate of Local Training           & 0.01    \\
    \hline
    Learning Rate Decay after Each Round              & 0.99    \\
    \hline
    Momentum of Local SGD                             & 0.9     \\
    \hline
    Weight Decay of Local SGD                         & 5$e^{-4}$ \\
    \hline
    Batch Size                                        & 10      \\
    \hline
    Non-IIDness                                       & \makecell[c]{Dirichlet with $\alpha = 0.5$ on MNIST, \\ Fashion MNIST and CIFAR-10, \\ and naturally non-IID with feature distribution \\ skew on FEMNIST and Sent140 } \\
    \hline
    Poisoning Data Ratio                              & 0.3     \\
    \hline
    Ratio of Attackers                                & 0.2     \\
    \bottomrule[1.0pt]
  \end{tabular}
\end{table*}

\subsection{Hyperparameters Specialized for \app}
\label{subsec-appendix-hyper-app}
In this section, hyperparameters involved with \app are listed, and some of them have been provided previously.
$M$ is $\frac{K}{2}$. 
$\check{M} = 0.1K$. 
$M^E = 0.05K$ on FEMNIST and otherwise $0.04K$, $E^{VI}$ and $E^{VT}$ higher than 270 and 30, respectively.
$E^{VI}$ is 270 on MNIST and Fashion MNIST, 360 on FEMNIST and Sent140 and 450 on CIFAR-10.
$E^{VT}$ is 30 on MNIST and Fashion MNIST, 40 on FEMNIST and Sent140 and 50 on CIFAR-10.
$S^{H}$ and $S^{L}$ are 256 and 64 on all datasets, respectively.

From Section \ref{subsec-approach-clustering}, $\check{K}$ is pre-determined through Gap statistic \cite{tibshirani2001estimating-gap}, which recommend an appropriate number of clusters for K-means.
To avoid the additional overhead caused by real-time computation, Gap statistic is applied on the received model updates in the first round in a layer-wise manner, with the searching range is limited in [2, 15]. 
After that, the recommend $\check{K}$ on each layer is averaged, and kept unchanged.
Specifically, $\check{K}$ is set to 11, 11, 12, 11, 11 on MNIST, Fashion MNIST, CIFAR-10, FEMNIST and Sent140, respectively.

From Section \ref{subsec-approach-vae}, 
Thus, the larger $T^V$ is, the more reliable the top-down election is. 
However, if $T^V$ is too large, the convergence of global model will be jeopardized.
From Figure \ref{pic-ddif}, differences between $\Delta \mathbf{w}^B$ shrinks faster on data with label distribution skew than that on data with feature distribution skew. 
Thus, we empirically set $T^V$ to $T/4$ on MNIST and Fashion MNIST, and $T/3$ on CIFAR-10 which is more complex than previous two.
For datasets with feature distribution skew, we set empirically set $T^V$ to $T/2$.

\begin{table*}[t]
  \renewcommand\arraystretch{0.88}
  \caption{Performances of several selected approaches on MNIST (CBA) with different $\alpha$ and Malicious Client Ratio (MCR). All values are percentages.}
  \label{tab-seperated-evaluation}
  \setlength\tabcolsep{3pt}
  \centering
  \begin{tabularx}{\linewidth}{>{\centering\arraybackslash}X|>{\centering\arraybackslash}X>{\centering\arraybackslash}X|>{\centering\arraybackslash}X>{\centering\arraybackslash}X|>{\centering\arraybackslash}X>{\centering\arraybackslash}X|>{\centering\arraybackslash}X>{\centering\arraybackslash}X|>{\centering\arraybackslash}X>{\centering\arraybackslash}X}
    \toprule[1.0pt]
    \multirow{2}{*}{\diagbox{$\alpha$}{\!\!\!\!MCR}}  & \multicolumn{2}{c|}{30\%} & \multicolumn{2}{c|}{25\%} & \multicolumn{2}{c|}{20\%} & \multicolumn{2}{c|}{15\%} & \multicolumn{2}{c}{10\%} \\
    \cmidrule{2-11}
                                              & BA         & MA          & BA          & MA         & BA          & MA         & BA          & MA         & BA          & MA         \\
    \midrule[1.0pt]
    \multicolumn{11}{c}{\textbf{\app}} \\
    \midrule[1.0pt]
    0.1                                       & 100.0      & 93.31       & 100.0       & 92.63      & 97.85       & 95.36      & 0.46        & 94.15      & 0.32        & 95.37      \\
    0.5                                       & 0.28       & 98.62       & 0.31        & 98.87      & 0.19        & 98.79      & 0.25        & 98.53      & 0.14        & 98.44      \\
    1.0                                       & 0.33       & 98.96       & 0.18        & 98.88      & 0.14        & 98.79      & 0.15        & 98.88      & 0.12        & 98.77      \\
    \midrule[1.0pt]
    \multicolumn{11}{c}{\textbf{Krum}} \\
    \midrule[1.0pt]
    0.1                                       & 100.0      & 95.83       & 100.0       & 95.80      & 99.94       & 97.23      & 99.40       & 97.41      & 0.49        & 98.19      \\
    0.5                                       & 100.0      & 98.61       & 99.97       & 98.74      & 99.98       & 98.80      & 0.23        & 98.89      & 0.15        & 98.91      \\
    1.0                                       & 99.99      & 98.82       & 99.99       & 98.85      & 99.94       & 98.86      & 0.26        & 99.04      & 0.15        & 98.97      \\
    \midrule[1.0pt]
    \multicolumn{11}{c}{\textbf{CRFL}} \\
    \midrule[1.0pt]
    0.1                                       & 100.0      & 97.88       & 99.71       & 97.99      & 98.93       & 97.86      & 91.97       & 97.86      & 91.49       & 97.59      \\
    0.5                                       & 99.98      & 98.40       & 100.0       & 98.43      & 99.93       & 98.45      & 99.07       & 98.41      & 99.17       & 98.45      \\
    1.0                                       & 99.97      & 98.55       & 99.52       & 98.53      & 99.90       & 98.52      & 99.21       & 98.44      & 99.19       & 98.56      \\
    \midrule[1.0pt]
    \multicolumn{11}{c}{\textbf{RLR}} \\
    \midrule[1.0pt]
    0.1                                       & 99.98      & 96.51       & 100.0       & 95.82      & 99.98       & 96.33      & 99.80       & 95.15      & 99.14       & 95.23      \\
    0.5                                       & 100.0      & 97.35       & 99.78       & 97.18      & 99.98       & 97.71      & 99.49       & 97.28      & 64.28       & 97.92      \\
    1.0                                       & 99.99      & 97.75       & 100.0       & 97.86      & 99.99       & 97.91      & 99.63       & 97.62      & 4.68        & 98.31      \\
    \midrule[1.0pt]
    \multicolumn{11}{c}{\textbf{DnC}} \\
    \midrule[1.0pt]
    0.1                                       & 99.92      & 95.57       & 99.85       & 95.85      & 99.37       & 96.05      & 100.0       & 96.92      & 90.02       & 96.26       \\
    0.5                                       & 99.99      & 98.81       &100.00       & 98.77      & 0.12        & 98.89      &  0.95       & 98.88      & 0.23        & 98.86       \\
    1.0                                       & 99.99      & 98.94       & 99.99       & 98.93      & 0.11        & 98.88      &  0.23       & 98.94      & 0.20        & 98.91       \\
    \midrule[1.0pt]
    \multicolumn{11}{c}{\textbf{FLAME}} \\
    \midrule[1.0pt]
    0.1                                       & 100.0      & 96.17       & 100.0       & 96.51      & 100.0       & 96.47      & 0.42        & 96.90      & 0.65        & 96.37      \\
    0.5                                       & 100.0      & 98.46       & 99.98       & 98.46      & 0.48        & 98.77      & 0.19        & 98.45      & 0.20        & 98.60      \\
    1.0                                       & 99.99      & 98.70       & 99.99       & 98.85      & 0.23        & 98.84      & 0.19        & 98.87      & 0.21        & 98.82      \\
    \midrule[1.0pt]
    \multicolumn{11}{c}{\textbf{FLDetector}} \\
    \midrule[1.0pt]
    0.1                                       &100.00      & 97.82       & 99.54       & 97.19      & 98.92       & 98.06      & 96.30       & 96.99      & 89.37       & 97.65       \\
    0.5                                       &100.00      & 98.95       & 99.99       & 98.96      & 100.0       & 98.84      & 99.97       & 98.97      & 99.92       & 98.95       \\
    1.0                                       &99.99       & 99.07       & 99.98       & 98.97      & 99.99       & 98.93      & 99.90       & 99.04      & 99.89       & 99.00       \\
    \bottomrule[1.0pt]
  \end{tabularx}
\end{table*}

\subsection{Hyperparameters Specialized for Comparison Approaches}
Hyperparameters in compared approaches are set as the recommended values in their corresponding papers or in their open-source implementation.
Here, we provide the detailed hyperparameters of compared approaches. 
Note that the hyperparameters of these approaches are denoted by characters in their corresponding papers, and some of them may appear repeated in this paper.
The specific meaning of each hyperparameter can refer to the corresponding paper.
The values of them are based on their corresponding papers.
For Krum, the ratio of estimated malicious clients is set to 0.3 to tolerant more malicious clients.
For CRFL, $\sigma$ is set to 0.01, $\rho$ is set to 15.0, $M$ (number of noised models) is set to 20.
For RLR, $\theta$ is 10 on MNIST, Fashion MNIST, CIFAR-10 and Sent140, and 20 on FEMNIST.
For FLAME, $\lambda$ is set to 0.001. 
For FLDetector, the window size $N$ is set to 5.
For DnC, the dimension of subsampling is set to 10,000.
For FLIP, we keep the hyperparameters the same as in its original paper on corresponding datasets. 

\section{Supplementary Experiments}
\label{sec-appendix-exp}

\subsection{Separated Evaluation on Non-IIDness and Ratio of Malicious Clients}
\label{subsec-appendix-separated}
We conduct experiments to fully investigate the effects of data heterogeneity and the percentage of malicious clients by evaluating the proposed method and some of the baselines with different combination of non-IIDness and ratio of malicious clients. 
These experiments are conducted on MNIST with CBA attacks as described in Section \ref{sec-exp}.

When the non-IIDness is relative complex, i.e., medium heterogeneous, which is relatively more common in real world, Snowball performs well to defend against a relatively large ratio of attackers. 
The above findings and analysis align with our claim in the paper that Snowball can effectively defend backdoor attacks with complex non-IIDness, a not high PDR and a relatively large attacker ratio.

But we can also find from Table \ref{tab-seperated-evaluation} that the resistance of Snowball to backdoor attacks will be negatively impacted by extremely heterogeneous data ($\alpha=0.1$) when the ratio of malicious clients is no less than 20\%. 
Actually, we focus on the data with complex non-IIDness instead of extremely heterogeneous data, while more severe data heterogeneity may be not as complex as the medium data heterogeneity. Taking this issue into an extreme example, in the most severe heterogeneity scenario, each client may contain data within only one category, where the data distribution is relatively simple. In this work, we focus on heterogeneity with more complex non-IIDness (medium heterogeneity or feature skew), which is relatively more common in real world.
As we have mentioned in Section \ref{sec-intro}, it may be difficult to have a one-size-fits-all approach.
Each method has its own strengths in particular scenarios. 
In real world, to defend against various attacks and keep the safety of the FL systems, it may be better to have several defenses integrated together. 
One advantage of \app is that it can be easily integrated into existing FL systems in a non-invasive manner, since it only filters out several model updates for aggregation.

Besides, more severe heterogeneity negatively affects MA of both Snowball and the baselines, because they are not personalized FL approaches.

\subsection{The selection of $M$}
$M$ decides how many model updates are aggregated during one round of federated aggregation. Intuitively, a larger $M$ will bring higher MA (main task accuracy), since the average of more model updates is stable than that of less model updates in terms of convergence. This phenomenon is more pronounced when the data is non-IID, because if too few model updates are aggregated, it will lead to significant differences in the global model's trajectory between two consecutive rounds, jeopardizing the accuracy of global model. 
However, a larger $M$ may cause infected model updates wrongly selected and aggregated, since as the top-down election gradually selects model updates, the proportion of infected model updates among the remaining candidates increases. Therefore, if $M$ is set too large, there is a higher likelihood that infected models will be mistakenly chosen during the final few steps of top-down election.

Thus, the value of $M$ is a tradeoff between MA and BA.
We set $M$ to $\frac{K}{2}$ based on the consideration that the number of malicious clients must not exceed the number of benign clients. Due to the significant consequence that higher BA may cause, we have opted for a relatively conservative strategy of $M$.

To experimentally demonstrate this, we run Snowball with different $M$ on Fashion MNIST with the other settings the same as in Table \ref{tab-appendix-M} of our manuscript. 

\begin{table}[t]
  \caption{The performance of \app with different of $M$ on MNIST (CBA)}
  \label{tab-appendix-M}
  \centering
  \begin{tabular}{c|c|c}
  \toprule[1.0pt]
      $M$ & BA & MA \\ 
      \midrule[1.0pt]
      0.2$K$ & 0.17 & 89.13 \\ 
      0.3$K$ & 0.23 & 89.21 \\ 
      0.4$K$ & 0.39 & 89.23 \\ 
      0.5$K$ & 0.39 & 89.27 \\ 
      0.6$K$ & 0.17 & 89.63 \\ 
      0.7$K$ & 0.22 & 90.07 \\ 
      0.8$K$ & 92.02 & 89.95 \\ 
  \bottomrule[1.0pt]
  \end{tabular}

\end{table}

From the results, we can find that MA increases with the increase of $M$, but if $M$ is set to 0.8$K$, the approach fails to defend against backdoor attacks, although there are 0.2$K$ malicious clients. Thus, $M$ should not be set excessively large in pursuit of higher MA.

\subsection{Time Consumption}
\label{subsec-appendix-exp-time}
\begin{figure*}[t]
  \centering
  \includegraphics[width=0.7\linewidth]{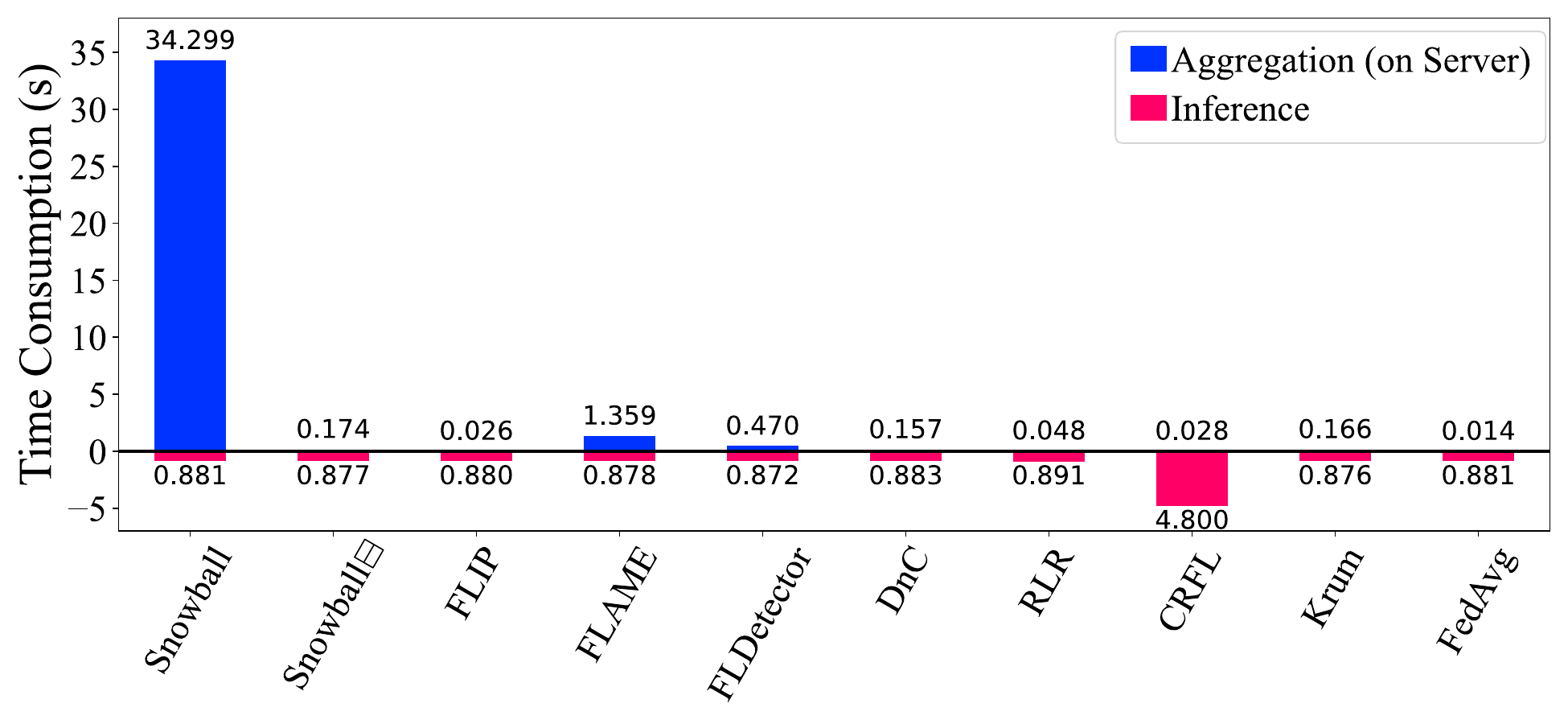}
  \caption{Time consumption of approaches, where the aggregation time is consumed by the server, and the inference time is calculated by conduct model inference on the test set.}
  \label{pic-appendix-time}
\end{figure*}
The main limitation of \app lies in its complexity mainly brought by the training, repeated tuning and prediction of the VAE.
Fortunately, such overhead is added only at the server, which is usually equipped with powerful computation resources supporting massive parallel computing. 
Besides, the aggregation in FL is not very frequent. 
In contrast, the main time consumption of FL usually comes from training and transmitting.

We run these approaches on MNIST at one of the previously mentioned platform, i.e., an Ubuntu 22.04.2 platform with an Intel(R) Core(TM) i9-12900K CPU, an NVIDIA RTX 3090 GPU and 64 GB RAM, to have a numerical result on their time consumption. 
Figure \ref{pic-appendix-time} presents the time consumption of these approaches taken by aggregation of model updates and inference on the test set, respectively.
As presented, \app takes more time than its peers and \appnoblank$\boxminus$ for aggregation.
\appnoblank$\boxminus$ does not incur significant additional time overhead compared with the peers of \appnoblank.
This indicates that the main consumption of \app lies in the training, repeated tuning and prediction of the VAE. 

For inference, we can find that \app and its peers except CRFL take approximately the same amount of time for inference on the test set, while CRFL takes significantly more time compared to other approaches, because it relies on model ensembles for inference. 

As previously mentioned, the additional time overhead brought by \app is added only at the server.
On one hand, the server usually has powerful computational resources, and by leveraging large-scale parallel computing, significant reductions in aggregation time can be achieved. 
On the other hand, the aggregation in FL does not typically occur frequently because local training tends to take longer time compared to the aggregation.

\end{document}